\documentclass{article}

\PassOptionsToPackage{table,dvipsnames}{xcolor}

\usepackage{PRIMEarxiv}
\usepackage[utf8]{inputenc}
\usepackage[T1]{fontenc}

\usepackage{hyperref}
\usepackage{url}
\usepackage{booktabs}
\usepackage{amsfonts}
\usepackage{amssymb}
\usepackage{amsmath}
\usepackage{nicefrac}
\usepackage{microtype}
\usepackage{graphicx}
\usepackage{multirow}
\usepackage{array}
\usepackage{tabularx}
\usepackage{siunitx}
\usepackage{listings}
\usepackage{enumitem}
\usepackage{tcolorbox}
\tcbuselibrary{listings,skins}
\usepackage[section]{placeins}   
\usepackage{tikz}
\usetikzlibrary{arrows.meta,positioning,fit,backgrounds}
\pgfdeclarelayer{behind}
\pgfsetlayers{behind,background,main}

\newcolumntype{C}{>{\centering\arraybackslash}X}



\definecolor{lstbg}{HTML}{F8FAFC}
\definecolor{lstkeyword}{HTML}{1E40AF}
\definecolor{lststring}{HTML}{0D9488}
\definecolor{lstcomment}{HTML}{64748B}
\definecolor{lstfunc}{HTML}{9333EA}
\definecolor{lsttitlebg}{HTML}{1E293B}
\definecolor{linkblue}{HTML}{1E40AF}
\definecolor{urlteal}{HTML}{0D9488}

\hypersetup{
    pdftitle={CircuitKIT: Circuit Discovery, Evaluation, and Application Toolkit for Mechanistic Interpretability},
    pdfauthor={Pratinav Seth, Hem Gosalia, Aditya Kasliwal, Vinay Kumar Sankarapu},
    pdfsubject={Mechanistic Interpretability Toolkit},
    pdfkeywords={mechanistic interpretability, circuit discovery, attribution patching,
                 model compression, faithfulness evaluation, custom-data tasks, source-available toolkit},
    colorlinks=true,
    linkcolor=linkblue,
    citecolor=linkblue,
    urlcolor=urlteal,
}

\setlist[itemize]{parsep=0pt, itemsep=3pt}
\setlist[enumerate]{parsep=0pt, itemsep=3pt}

\newcommand{\CK}{\emph{CircuitKIT}}

\newcommand\blfootnote[1]{{\begingroup\renewcommand\thefootnote{}\footnote{#1}\addtocounter{footnote}{-1}\endgroup}}

\lstdefinelanguage{yaml}{
  morekeywords={true,false,null,y,n},
  sensitive=false,
  morecomment=[l]{\#},
  morestring=[b]",
  morestring=[b]',
  alsoletter={-},
  moredelim=[l][\color{black}\ttfamily]{-},
}

\newtcblisting{codeblock}[1][]{
    enhanced,
    listing only,
    colback=lstbg,
    colframe=lsttitlebg,
    boxrule=0.25pt,
    arc=3pt,
    outer arc=3pt,
    top=4pt, bottom=2pt, left=0pt, right=0pt,
    before skip=4pt,
    after skip=4pt,
    title=\hspace{2pt},
    fonttitle=\ttfamily\tiny\color{lstcomment},
    coltitle=lstcomment,
    colbacktitle=lsttitlebg,
    attach boxed title to top left={yshift=0pt, xshift=0pt},
    boxed title style={
        size=minimal,
        height=5pt,
        width=\textwidth+0.5pt,
        arc=3pt,
        bottom=0pt,
        colback=lsttitlebg,
        boxrule=0pt,
    },
    listing options={
        basicstyle=\footnotesize\ttfamily\color{black!80},
        keywordstyle=\color{lstkeyword}\bfseries,
        commentstyle=\color{lstcomment}\itshape,
        stringstyle=\color{lststring},
        identifierstyle=\color{black},
        emph={Pipeline,Circuit,CircuitScores,discover,evaluate,prune,quantize,export,summary,
              visualize,from_custom_data,load_model,faithfulness,export_checkpoint,benchmark,
              selective_finetune,visualize_circuit,load_scores,template_normalize,
              clean_only_normalize,ActivationSteering,RomeHandler,MemitHandler,CircuitLoRA,
              HallucinationDetector,ComparisonDashboard,get_covariance,register,
              auto_task_from_hf,run_full_faithfulness,discover_circuit,TaskSpec},
        emphstyle=\color{lstfunc}\bfseries,
        numbers=none,
        breaklines=true,
        breakatwhitespace=true,
        showstringspaces=false,
        tabsize=4,
        xleftmargin=0.3em,
        columns=flexible,
        keepspaces=true,
        lineskip=0.5pt,
        aboveskip=2pt,
        belowskip=2pt,
    },
    #1,
}

\newtcolorbox{takeawaybox}[1][Takeaway]{%
  colback=black!4!white, colframe=black!60!white,
  fonttitle=\bfseries\footnotesize, colbacktitle=black!80!white,
  coltitle=white, enhanced,
  attach boxed title to top left={yshift=-2.2mm, xshift=4mm},
  boxrule=0.6pt, arc=1.5pt, left=5pt, right=5pt, top=6pt, bottom=3pt,
  title={#1}}

\title{\includegraphics[width=0.69\linewidth]{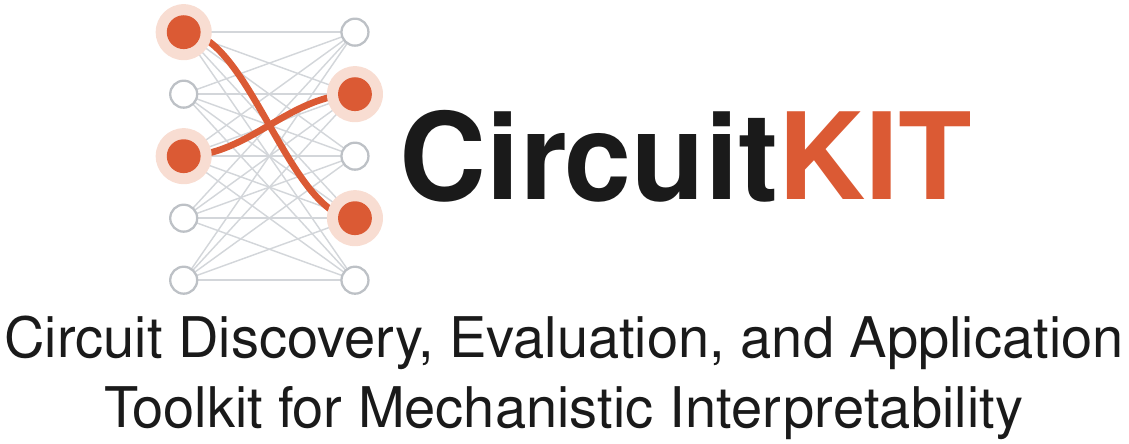}}

\author{
  Pratinav Seth$^{*}$, Hem Gosalia$^{*\dagger}$, Aditya Kasliwal$^{*\dagger}$, \\Vinay Kumar Sankarapu \\
  \affiliation{Lexsi Labs}\\
  \texttt{pratinav.seth@lexsi.ai}
}

\setabstract{%
Circuit analysis can support not only model explanation but also downstream interventions such as pruning, editing, steering, and selective fine-tuning. However, conducting such analyses currently requires stitching together separate implementations for discovery, evaluation, and intervention, as well as hand-authoring the contrastive prompts required by many discovery methods. This fragmentation makes methods difficult to compare and limits their application beyond canonical tasks. We introduce \emph{CircuitKIT}, a source-available library that connects the circuit-analysis workflow through a typed, serializable representation. \emph{CircuitKIT} provides a suite of discovery algorithms, declarative interfaces for mapping structured data into discovery tasks, complementary circuit diagnostics, and downstream application modules. Together, these components provide common infrastructure for conducting and comparing circuit analyses. The library, examples, notebooks, and documentation are released at \url{https://github.com/Lexsi-Labs/CircuitKIT}.
}

\setkeywords{mechanistic interpretability, automated circuit discovery, transformer circuits, circuit faithfulness, actionable interpretability, large language models, interpretability toolkit}

\runningtitle{CircuitKIT: Circuit Discovery, Evaluation, and Application Toolkit for Mechanistic Interpretability}

\begin{document}
\maketitle
\blfootnote{$^{*}$Equal contribution.\quad $^{\dagger}$Work done while at Lexsi Labs.}
\vspace{-0.35cm}   

\section{Introduction}
\label{sec:introduction}

Circuit analysis is one of the main approaches to mechanistic interpretability of large language models~\cite{elhage2021mathematical, wang2023interpretability, conmy2023towards}. The workflow decomposes into three stages: \emph{discover} a circuit, the subgraph of attention heads, MLP sublayers, or individual neurons causally responsible for a target behavior; \emph{evaluate} whether the discovered circuit is faithful, stable, and robust; and \emph{intervene} on the model using the circuit, for example to prune, quantize, edit, or fine-tune it. Each stage has accumulated multiple competing methods: a growing family of discovery algorithms~\cite{syed-etal-2024-attribution, hanna2024faith, zhang2025eapgp, conmy2023towards, bian2025ibcircuit, hsu2025cdt}, several evaluation criteria~\cite{mueller2025mib, gupta2024interpbench, miller2024faithfulness}, and an expanding set of downstream interventions~\cite{meng2022rome, ma2023llmpruner, turner2025steering}. These methods ship in separate per-paper repositories with incompatible artifact formats. A practitioner, whether an interpretability researcher comparing discovery methods on their own task or an engineer who wants a compression or safety audit grounded in a model's actual mechanism, must currently stitch together several codebases and write custom glue for every pairwise combination of discovery method, evaluation criterion, and intervention. The resulting pipelines are difficult to reproduce, difficult to compare across methods, and expensive to extend to new tasks.

There is a second, quieter cost to this fragmentation. Faithfulness scores depend on the ablation method used to compute them, and switching that one choice can \emph{reverse} the ranking of discovery methods~\cite{miller2024faithfulness}; because each repository hard-codes its own choice, numbers do not compare across papers, and every method--criterion--intervention combination needs fresh glue code before it can be compared at all.

Existing higher-level toolkits cover only part of this pipeline. TransformerLens~\cite{nanda2022transformerlens} and NNsight~\cite{fiottokaufman2024nnsight} provide framework-level introspection substrates that \emph{can} express all three stages programmatically, but ship no standardized task interface, no multi-criterion evaluation framework, and no downstream-intervention contract. Auto-Circuit~\cite{miller2024faithfulness} bundles gradient-attribution discovery with patching evaluation for a fixed task set at GPT-2 scale. MIB~\cite{mueller2025mib} and InterpBench~\cite{gupta2024interpbench} are fixed-dataset evaluation benchmarks. None spans the full discover, evaluate, and intervene pipeline on multi-billion-parameter language models with a single shared artifact (Table~\ref{tab:library-comparison}, Section~\ref{sec:related}).

A finer gap lies on the data side. Attribution-based discovery methods such as EAP~\cite{syed-etal-2024-attribution} and EAP-IG~\cite{hanna2024faith} are intrinsically contrastive: every example must arrive as a (clean, corrupt) prompt pair so the algorithm can attribute the model's differential behavior to individual components. For canonical tasks like Indirect Object Identification (IOI)~\cite{wang2023interpretability} this pairing is freely available. For structured user data (a CSV of question-answer pairs, an MCQ benchmark, or a HuggingFace classification split), constructing well-formed pairs is the practical bottleneck that has kept circuit discovery largely confined to the canonical task set.

\emph{CircuitKIT} addresses both gaps: it composes the full discover, evaluate, and intervene pipeline behind one shared artifact on multi-billion-parameter language models, and maps structured user datasets into circuit-discovery tasks through a declarative template interface (Table~\ref{tab:library-comparison}).

\paragraph{Contributions.}
This paper makes three systems contributions:

\begin{enumerate}
    \item \emph{CircuitKIT, an integrated library for circuit analysis.}
    We introduce and release a source-available library that brings thirteen circuit-discovery algorithms across four backend families into a common workflow. A typed, serializable \texttt{CircuitScores} artifact connects these implementations and makes their outputs accessible through a stateful pipeline, a functional API, and a YAML-driven CLI (Sections~\ref{sec:architecture}--\ref{sec:applications}, \ref{sec:usage}).

    \item \emph{A declarative path from structured data to circuit discovery.}
    \emph{CircuitKIT} maps datasets into discovery tasks, handling various aspects. This enables circuit discovery on user-defined data without requiring a bespoke task implementation for each dataset (Section~\ref{sec:tasks}).

    \item \emph{A common layer for evaluating and applying discovered circuits.}
    \emph{CircuitKIT} connects discovery outputs to six complementary diagnostics, matched baselines, and seven downstream application modules. Checkpoint export and standardized benchmark integration allow circuit hypotheses to be assessed both intrinsically and through their downstream effects under a common experimental interface (Sections~\ref{sec:evaluation}--\ref{sec:extensibility}).
\end{enumerate}



\section{Background and Related Work}
\label{sec:background}
\label{sec:related}

\emph{CircuitKIT} operates on the vocabulary of mechanistic interpretability, which we review here for readers new to it; those already versed in circuit analysis can skip to Section~\ref{sec:architecture}. 

A decoder-only transformer routes every token position through a shared \emph{residual stream} that each attention head and MLP sublayer reads from and writes back to, so the model can be read as an addressable collection of components (attention heads, MLP sublayers, and their individual neurons) wired into a \emph{computational graph} whose edges are the paths by which one component's output feeds another~\cite{elhage2021mathematical}. 
A \emph{circuit} is the subgraph causally responsible for a specific behavior: the small mechanism the model actually uses, while the rest of the network stays largely irrelevant~\cite{wang2023interpretability}.

\emph{Circuit discovery} locates that subgraph by causal intervention. \emph{Activation patching} runs the model on a \emph{clean} input, overwrites a component's activation with the value it takes on a contrasting \emph{corrupt} input, and scores importance by how far the output moves, so a behavior must be posed as a \emph{(clean, corrupt)} pair whose \emph{logit difference}, for a canonical task such as Indirect Object Identification, isolates the responsible components~\cite{wang2023interpretability, meng2022rome, hanna2023greaterthan}. Because patching every edge separately is costly, \emph{attribution patching} approximates all edges in a single forward-backward pass through a first-order, optionally integrated-gradient estimate~\cite{syed-etal-2024-attribution, hanna2024faith, sundararajan2017axiomatic}, whereas search-based methods such as ACDC prune edges without any linear approximation~\cite{conmy2023towards}. 

A discovered circuit is finally judged for \emph{faithfulness}: whether it reproduces the behavior once every out-of-circuit component is \emph{ablated} (replaced by a zero, mean, or resampled value). Since that verdict depends on which ablation is used and can even reverse method rankings~\cite{miller2024faithfulness}, \emph{CircuitKIT} reports a panel of complementary checks rather than a single number (Section~\ref{sec:evaluation}).

\subsection{Related Work}
\begin{table}[pt]
\centering
\caption{Coverage across the three pipeline stages. \emph{Discovery} counts stable discovery algorithms available (of 13 total); \emph{Evaluation} counts supported evaluation diagnostics out of six; \emph{Application} counts application modules that consume the shared artifact. ``$+{\sim}n$'' indicates partial or experimental support. Only \emph{CircuitKIT} spans all three stages behind one artifact with a template-driven custom-data path. \emph{CircuitKIT} builds on the TransformerLens~2.x series (3.x support is planned, Section~\ref{subsec:coverage}).}
\vspace{2mm}
\label{tab:library-comparison}
\footnotesize
\begin{tabular}{@{}lcccc@{}}
\toprule
Library & Discovery & Evaluation\ /6 & Application & Custom Data \\
\midrule
TransformerLens & 0 & 0 & 0 & No \\
NNsight / NDIF & $0{+}{\sim}1$ & $+{\sim}1$ & 0 & No \\
Captum & $+{\sim}1$ & $+{\sim}1$ & 0 & No \\
Auto-Circuit & 2 & $+{\sim}2$ & 0 & No \\
MIB & 2 & 2 & 0 & No \\
InterpBench & 0 & $+{\sim}1$ & 0 & No \\
\textbf{\emph{CircuitKIT}} & \textbf{6 (of 13)} & \textbf{6} & \textbf{7} & \textbf{Yes} \\
\bottomrule
\end{tabular}
\end{table}
\paragraph{Model introspection frameworks.} TransformerLens~\cite{nanda2022transformerlens} and NNsight/NDIF~\cite{fiottokaufman2024nnsight} give structured access to model internals through hook-based and context-manager APIs, and both can express circuit discovery, evaluation, and intervention programmatically across many HuggingFace architectures. Neither, however, ships a standardized task interface, a multi-pillar evaluation framework, or a downstream-intervention contract over a shared artifact. Captum~\cite{kokhlikyan2020captum} offers generic gradient-attribution for PyTorch models but is not specialized for circuit-level mechanistic interpretability. \emph{CircuitKIT} builds on TransformerLens as its loading and hooking substrate and adds exactly the task, evaluation, and intervention layers those substrates leave to the user.

\paragraph{Single-stage toolkits.} Auto-Circuit~\cite{autocircuit2023, miller2024faithfulness} bundles EAP and EAP-IG discovery with patching evaluation for a fixed task set at GPT-2 scale, but does not extend to evaluation criteria beyond patching or to downstream interventions. MIB~\cite{mueller2025mib} and InterpBench~\cite{gupta2024interpbench} are fixed-dataset benchmarks for ranking discovery methods on their included tasks; neither is extensible to user data nor connected to an intervention layer. \emph{CircuitKIT} differs in spanning all three stages behind one artifact and in providing declarative adapters for structured user data.

\paragraph{Custom-data and contrastive pairing.} None of the toolkits above synthesizes the corrupt half of a contrastive pair from a clean prompt; all assume the user supplies both halves. The closest concurrent work generates synthetic \emph{templates} for attribution patching~\cite{templategen2025}, producing prompts to discover on; \emph{CircuitKIT} instead operates over the user's own examples, which is the case that arises when the behavior of interest is defined by data the user already has and cannot be re-synthesized without changing what is being studied. \emph{CircuitKIT}'s template-driven path (Section~\ref{subsec:custom-data}) closes that gap by making pairing, alignment, and validation a declarative step in the same pipeline, and by routing clean-only data to IBCircuit and CD-T when no counterfactual exists at all.

\paragraph{Attribution and stability analysis.} Beyond EAP and EAP-IG, recent methods such as EAP-GP~\cite{zhang2025eapgp} and RelP~\cite{relp2025} offer alternative approximations to activation patching; \emph{CircuitKIT} ships EAP-GP at stable tier and RelP at research tier, both reachable through the selector registry (Section~\ref{sec:extensibility}). On the evaluation side, Miller et al.~\cite{miller2024faithfulness} show that single faithfulness metrics depend on the ablation method and can reverse rankings, which directly motivates the multi-pillar design; circuit-restricted weight-surgery work~\cite{kasliwal2026cdeltatheta} intervenes through discovered circuits but reports a single discovered circuit and so cannot expose discovery-induced variance, which the seventh pillar is built to measure.

\paragraph{Interventions and actionable interpretability.} A growing line of work turns localized structure into concrete model changes: mechanistic localization guides robust knowledge unlearning and editing~\cite{guo2024mechunlearning}, activation additions steer behavior at inference time~\cite{turner2025steering}, and rank-one updates rewrite factual associations~\cite{meng2022rome, meng2023memit}. Circuit-scoped edits are increasingly used for safety: circuit-restricted weight arithmetic removes refusal behavior offline through a discovered circuit~\cite{kasliwal2026cdeltatheta}, circuit attribution can make unlearning persist through post-training quantization~\cite{sadhu2026forgetting}, and causal audits of neuron selectors test whether attribution actually localizes the components responsible for a behavior~\cite{eswar2026refusal}. 

Recent work frames this locate-then-act loop as \emph{actionable} interpretability~\cite{zhang2026actionable} and argues for interpretability as a design principle for alignment~\cite{sengupta2025interpalign}. These methods typically ship as standalone codebases tied to one intervention and one discovery choice. \emph{CircuitKIT} instead lets any of them consume the same discovered circuit and scores the result extrinsically (Section~\ref{sec:applications}), so the discovery method and the intervention that acts on it are decoupled rather than baked into one pipeline.

Table~\ref{tab:library-comparison} places \emph{CircuitKIT} against the closest existing frameworks. Four properties set it apart: a single typed artifact that discovery produces and later stages consume, so the discovery method, task, or intervention is selected through a common configuration interface rather than new glue code; synthesis of the corrupt half of a contrastive pair when the user has only clean data, rather than an assumption that both halves are supplied; a faithfulness \emph{panel} of complementary checks rather than a single score, which surfaces the known dependence of faithfulness on the ablation method; and an explicit split between stable and research tiers, so a reader can tell which methods carry cross-model evidence and a contributor has a clear path.

\begin{figure}[pt]
    \centering
    \includegraphics[width=\linewidth]{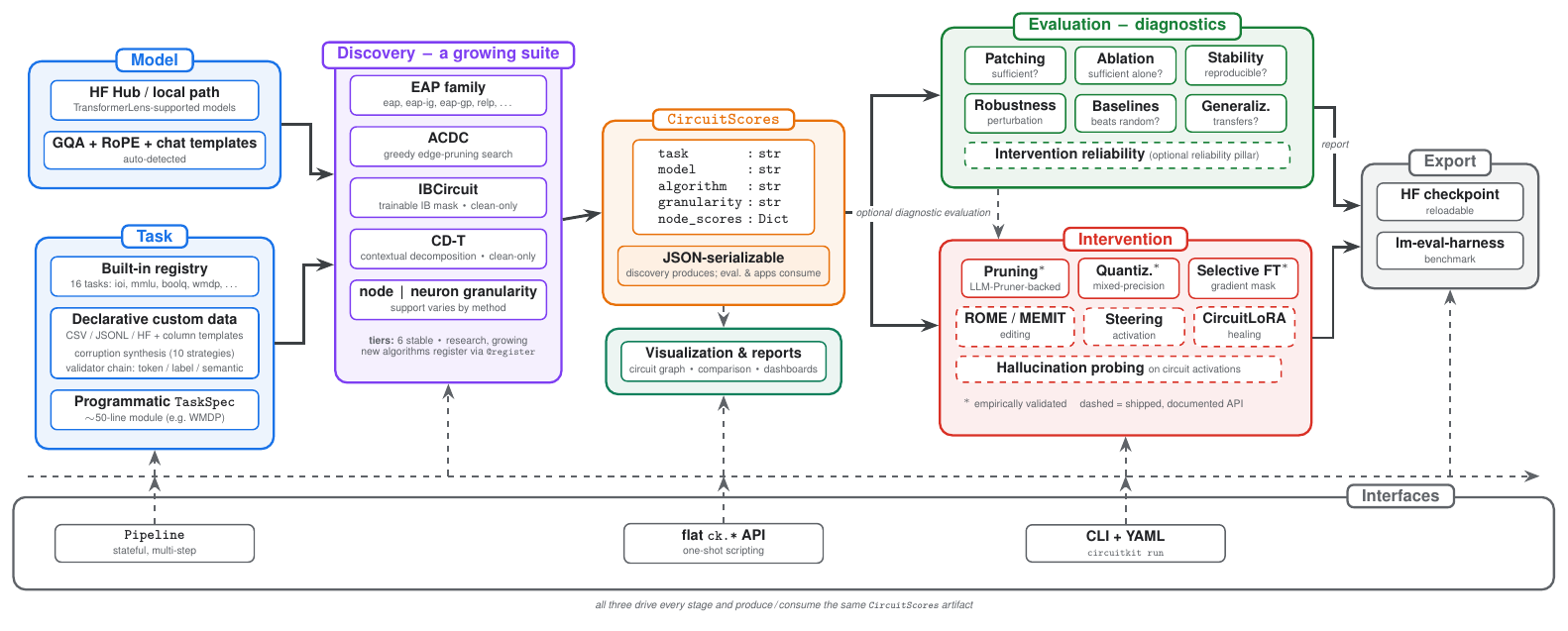}
    \caption{\emph{CircuitKIT} pipeline. A model and a task specification
    (built-in registry, template-driven custom data, or programmatic
    \texttt{TaskSpec}) feed a growing suite of discovery algorithms across four
    backends (six stable-tier shown, more registering as the field moves). Every discovery backend produces the same typed
    \texttt{CircuitScores} artifact, which is consumed by the evaluation
    diagnostics and downstream application modules; interventions export a reloadable
    HuggingFace checkpoint benchmarked via \texttt{lm-evaluation-harness}.
    $^{*}$~marks the three modules studied in
    Section~\ref{sec:experiments}; dashed pills ship in the library
    (Section~\ref{sec:applications}). The three user interfaces invoke the same
    underlying workflow and artifact contract.}
    \label{fig:pipeline}
\end{figure}
\section{Architecture}
\label{sec:architecture}

\emph{CircuitKIT} is a source-available Python library (Lexsi Labs Source Available License, LSAL~v1.1; see Section~\ref{sec:ethics}) built on TransformerLens~\cite{nanda2022transformerlens}. The library composes four stages (model loading, circuit discovery, faithfulness evaluation, and downstream intervention) behind a single shared artifact, \texttt{CircuitScores}: a typed, serializable record carrying the task descriptor, model identifier, discovery algorithm, granularity level (node or neuron), per-component attribution scores, and the discovery configuration. Discovery backends produce \texttt{CircuitScores}; evaluation and application modules consume it through standardized interfaces. Evaluation additionally produces a \texttt{FaithfulnessReport}, while application modules produce modified models, checkpoints, or diagnostic outputs (Figure~\ref{fig:pipeline}).

The design follows the pattern that HuggingFace's \texttt{Trainer}~\cite{wolf-etal-2020-transformers} established for training loops: a single object owns the workflow, sensible defaults cover the common case, and every stage remains individually addressable. Model to evaluated circuit is four lines:

\begin{codeblock}[]
from circuitkit import Pipeline

pipe = Pipeline("meta-llama/Llama-3.2-1B", task="ioi")  # or "gpt2" for a quick start
pipe.discover(algorithm="eap-ig", level="node")
report = pipe.evaluate()          # six pillars, one call
\end{codeblock}

\medskip
\noindent Each interface operates over the same artifact contract, allowing a circuit discovered through one interface to be evaluated or applied through another without format conversion: the same workflow runs as stateless function calls (the flat \texttt{ck.*} API) or a version-controlled YAML file (the CLI), and extends to a pruned, exported, harness-scored checkpoint with three more calls; Section~\ref{sec:usage} shows all three interfaces end to end.

\clearpage
\section{Circuit Discovery}
\label{sec:discovery}

\emph{CircuitKIT} ships thirteen discovery algorithms across four backend families, organized into explicit stability tiers that communicate how much cross-model validation each has received. Six are stable tier and have been tested across the GPT-2, Llama, Gemma, and Qwen families; the remaining seven are research tier, implemented and validated on GPT-2/IOI but not yet exercised at scale or across architectures. A single dispatch point selects among them by name, so the same discovery call reaches any backend:

\begin{codeblock}[]
import circuitkit as ck

model   = ck.load_model("meta-llama/Llama-3.2-1B")
circuit = ck.discover(model, task="ioi", algorithm="eap-ig",
                      level="neuron", sparsity=0.3,
                      output_path="./circuit.pt")
\end{codeblock}

\subsection{Model Coverage}
\label{subsec:coverage}

Model support has two layers, which we state separately because they carry different guarantees.

For \emph{discovery and evaluation}, \emph{CircuitKIT} loads models through TransformerLens (verified against the 2.18 release series, with TransformerLens~3 support planned for the immediate future), so circuit discovery and all evaluation pillars run on any model TransformerLens can load: GPT-2, Pythia, GPT-Neo, Llama, Gemma, Qwen, Mistral, Phi, Falcon, and the rest of its catalog of more than two hundred architectures, including models with grouped-query attention and rotary position embeddings. Chat templates for instruction-tuned checkpoints are auto-detected. The cross-model study in Section~\ref{subsec:e2} exercises this layer across six families.

For \emph{interventions}, the application modules additionally require an architecture-registry entry: a declarative mapping from component names to the concrete weight matrices needed to prune heads, assign quantization tiers, or mask gradients. Entries currently ship for the Llama, Qwen, and Gemma families (including Gemma~3), the families used throughout Section~\ref{sec:experiments}, with structurally complete entries for Mistral, Phi, GPT-2, and Falcon. Supporting a new family requires implementing only this registry interface, exactly as the three primary families were configured, with no changes to any discovery backend or evaluation pillar; Section~\ref{sec:extensibility} describes the registry pattern, and broader built-in coverage is planned as future work.

This two-layer design is deliberate. Discovery and evaluation operate on activations through hooks, which TransformerLens standardizes across architectures, so they generalize for free. Interventions rewrite weights, whose naming and layout are architecture-specific, so they are gated behind an explicit, auditable registry rather than best-effort string matching.

\subsection{The Stable Algorithms}
\label{subsec:stable-algos}

The stable tier spans complementary methodological approaches, gradient attribution, search-based pruning, information-bottleneck optimization, and contextual decomposition, chosen so that a user can cross-check a circuit found one way against a circuit found another on the same task without leaving the library.

\paragraph{Gradient-attribution methods (EAP, EAP-IG, EAP-GP).} Edge Attribution Patching (EAP)~\cite{syed-etal-2024-attribution} estimates the importance of each edge $(u \to v)$ in the model's computational graph by a first-order approximation to the effect of patching that edge from its clean to its corrupt activation. Writing $x_u$ and $x'_u$ for the clean and corrupt activations at the source of the edge and $\mathcal{L}$ for the task metric, the attribution score is
\begin{equation}
    s(u \to v) = (x_u - x'_u)^\top \left.\frac{\partial \mathcal{L}}{\partial x_v}\right|_{x_v = x_v^{\text{clean}}},
    \label{eq:eap}
\end{equation}
computed for all edges in a single forward-backward pass. EAP-IG~\cite{hanna2024faith} replaces the single-point gradient with an integrated gradient along the straight-line path from the corrupt to the clean activation,
\begin{equation}
    s_{\text{IG}}(u \to v) = (x_u - x'_u)^\top \frac{1}{m}\sum_{k=1}^{m} \left.\frac{\partial \mathcal{L}}{\partial x_v}\right|_{x_v = x'_v + \frac{k}{m}(x_v^{\text{clean}} - x'_v)},
    \label{eq:eapig}
\end{equation}
which satisfies the completeness and sensitivity axioms of Integrated Gradients~\cite{sundararajan2017axiomatic} at the cost of $m$ interpolation steps. EAP-GP~\cite{zhang2025eapgp} keeps the same activation-difference-times-gradient form but replaces the straight-line interpolation with an adaptively constructed integration path: starting at the clean embedding, each step descends the gradient of the squared output distance between the current point and the corrupt input, taking unit-norm steps, so the path follows the model's own geometry rather than a straight line. This costs roughly twice as much as EAP-IG at the same step count and can sharpen attribution where the straight-line path crosses regions of saturated gradient. All three are fast, scale to multi-billion-parameter models, and produce highly stable circuits across data resamples (Section~\ref{subsec:e1}). EAP-IG is the recommended default.

\paragraph{Automatic circuit discovery (ACDC).} ACDC~\cite{conmy2023towards} takes a search-based rather than an attribution-based approach: it greedily prunes edges from the full computational graph, removing an edge whenever doing so leaves the task metric within a tolerance of its full-model value, and keeps whatever survives. This recovers circuits without a linear approximation, at the cost of one forward pass per candidate edge, which makes it substantially more expensive than the EAP family. Because ACDC's decision unit is the edge between two named components, it operates at node granularity by construction (Section~\ref{subsec:granularity}).

\paragraph{Trainable information-bottleneck discovery (IBCircuit).} IBCircuit~\cite{bian2025ibcircuit} learns a soft mask over model components by gradient descent on a composite objective,
\begin{equation}
    \mathcal{L} = \alpha\,\mathcal{L}_{\text{task}} + \beta\,\mathcal{L}_{\text{IB}},
    \label{eq:ib-loss}
\end{equation}
where $\mathcal{L}_{\text{task}}$ preserves task performance under the masked model and $\mathcal{L}_{\text{IB}}$ is an information-bottleneck regularizer that drives the mask toward sparsity. The weights $\alpha$ and $\beta$ trade off fidelity against compactness. Because the objective depends only on the model's behavior on clean inputs, IBCircuit needs no corrupt counterpart, which makes it directly applicable when contrastive data is unavailable. It defaults to neuron-level masks and trains for a configurable number of epochs (1{,}000 by default).

\paragraph{Contextual decomposition (CD-T).} CD-T~\cite{hsu2025cdt} is a gradient-free, forward-pass method that splits every activation into a relevant and an irrelevant part and propagates the $(\text{rel}, \text{irrel})$ decomposition through each transformer component, attributing the prediction to source components by how much relevant signal they contribute. Like IBCircuit, it needs only clean inputs. \emph{CircuitKIT}'s implementation uses TransformerLens's unified weight API so the same propagation runs across GPT-2, Llama, Mistral, Gemma, Qwen, Falcon, and Pythia; its RoPE handling and gated-MLP cross-term split are documented approximations.

\begin{table}[!htbp]
\centering
\caption{Discovery algorithms shipped in \emph{CircuitKIT}, organized by backend family and stability tier. \emph{Stable}: tested across GPT-2, Llama, Gemma, and Qwen. \emph{Research}: implemented and validated on GPT-2/IOI, not yet exercised at scale or across architectures. ``Pairing'' indicates whether the algorithm requires a contrastive (clean, corrupt) pair or runs on clean inputs alone.}
\label{tab:discovery-algorithms}
\small
\begin{tabular}{@{}llcc@{}}
\toprule
Algorithm & Backend & Tier & Pairing \\
\midrule
\texttt{eap-ig}    & EAP       & Stable   & Contrastive \\
\texttt{eap}       & EAP       & Stable   & Contrastive \\
\texttt{eap-gp}    & EAP       & Stable   & Contrastive \\
\texttt{acdc}      & ACDC      & Stable   & Contrastive \\
\texttt{ibcircuit} & IBCircuit & Stable   & Clean-only \\
\texttt{cdt}       & CD-T      & Stable   & Clean-only \\
\midrule
\texttt{eap-ig-activations}  & EAP & Research & Contrastive \\
\texttt{eap-clean-corrupted} & EAP & Research & Contrastive \\
\texttt{eap-exact}           & EAP & Research & Contrastive \\
\texttt{atp-gd}              & EAP & Research & Contrastive \\
\texttt{relp}                & EAP & Research & Contrastive \\
\texttt{peap}                & EAP & Research & Contrastive \\
\texttt{eap-ifr}             & EAP & Research & Contrastive \\
\bottomrule
\end{tabular}
\end{table}

The stable tier reflects a coverage choice: gradient-attribution methods are cheap and scale well but rest on a local linear approximation; the search-based (ACDC) and optimization-based (IBCircuit) methods relax that approximation at higher cost; and the clean-only methods (IBCircuit, CD-T) remove the contrastive-data requirement entirely. Section~\ref{subsec:e1} reports the cross-algorithm comparison this makes possible. The research-tier variants (Table~\ref{tab:discovery-algorithms}) extend the EAP family with alternative integration paths and relevance-propagation schemes; they are reachable by name and governed by the same registry (Section~\ref{sec:extensibility}), so they can be promoted as validation evidence accumulates.

\subsection{Node and Neuron Granularity}
\label{subsec:granularity}

A distinguishing property of \emph{CircuitKIT} is that discovery runs at two granularities, selected by a single \texttt{level} argument, and \emph{both} are first-class rather than one being an aggregate of the other. At \emph{node} level, an entire attention head or MLP sublayer receives one importance score. At \emph{neuron} level, scores are computed for each attention-head channel and each MLP neuron, yielding a much finer circuit at higher discovery cost. Neuron-level scores matter because the intervention modules act on individual head channels and MLP neuron indices (Section~\ref{sec:applications}): a neuron-level circuit maps directly onto the units those modules prune, quantize, or fine-tune, with no lossy step down from a coarser node-level score. Every neuron-level study in Section~\ref{sec:experiments} relies on this correspondence.

\emph{Where in the model the scores are computed} is worth stating precisely, because it determines what a ``neuron'' means. For attention, both the EAP family and IBCircuit score at the per-head output (\texttt{attn.hook\_result} / \texttt{hook\_z}), so an attention neuron is a single head-dimension channel. For the MLP sublayer, there are two natural hook points, and \emph{CircuitKIT} exposes both through an \texttt{mlp\_hook} option:

\begin{itemize}
  \item \texttt{mlp\_out} (the default): the MLP is scored at its residual-stream interface (\texttt{hook\_mlp\_in} and \texttt{hook\_mlp\_out}), so an MLP neuron is one of the $d_{\text{model}}$ residual dimensions the sublayer reads and writes. This is the default because it keeps attention and MLP components on the same $d_{\text{model}}$ footing, which is what the residual-stream view of a circuit assumes, and it is the cheaper of the two.
  \item \texttt{post\_act}: the MLP is scored at its post-activation hidden layer (\texttt{mlp.hook\_post}), so an MLP neuron is one of the $d_{\text{mlp}}$ hidden units, the object most mechanistic-interpretability work means by ``MLP neuron.'' This is the more granular option (typically $d_{\text{mlp}} = 4 d_{\text{model}}$) and is selected when the goal is to localize computation to specific hidden units.
\end{itemize}

\noindent Both settings are available for the EAP family, IBCircuit, and CD-T; ACDC is node-only by construction, since its search operates over edges between named components and has no per-channel decision to make. Selecting the finer MLP hook is one field:

\begin{codeblock}[]
# Neuron-level discovery over the MLP hidden layer (d_mlp units)
circuit = ck.discover(model, task="ioi", algorithm="eap-ig",
                      level="neuron", mlp_hook="post_act",
                      output_path="./circuit_neuron.pt")
\end{codeblock}

\section{Task Specification and the Custom-Data Path}
\label{sec:tasks}

A target behavior is specified independently of how the circuit is found. Three task modes all feed the same discovery, evaluation, and intervention layers: a built-in registry, a template-driven custom-data path, and a programmatic protocol.

\subsection{Built-in Tasks}
\label{subsec:builtin-tasks}

Sixteen tasks are registered and accessed by a short key: \texttt{ioi}, \texttt{greater\_than}, \texttt{sva}, \texttt{hypernymy}, \texttt{gender\_bias}, \texttt{capital\_country}, \texttt{mmlu}, \texttt{glue} (SST-2 by default), \texttt{double\_io}, \texttt{wmdp}, \texttt{boolq}, \texttt{winogrande}, \texttt{winogrande\_mc}, \texttt{truthfulqa}, \texttt{ifeval}, and \texttt{gsm8k}. The library supplies prompts, metric functions, and dataloaders for each. Fifteen support discovery; \texttt{ifeval} is registered for downstream evaluation only, since instruction-following compliance has no natural contrastive pair. Passing a key is all that a built-in task requires:

\begin{codeblock}[]
pipe = Pipeline("meta-llama/Llama-3.2-1B", task="ioi")
pipe.discover(algorithm="eap-ig", level="neuron", sparsity=0.3)
\end{codeblock}

\subsection{The Template-Driven Custom-Data Path}
\label{subsec:custom-data}

The central data contribution of \emph{CircuitKIT} is that a compatible structured dataset can be mapped into a circuit-discovery task by declaring \emph{templates} over its columns, with no bespoke pairing code. The user supplies a CSV, JSONL, or HuggingFace dataset and writes a clean prompt template, a clean answer template, and, for contrastive algorithms, their corrupt counterparts. Placeholders in the templates name columns of the dataset; the pipeline fills them row by row, tokenizes, enforces token-length alignment between the clean and corrupt halves, precomputes the discriminative label, and hands a ready dataloader to the discovery backend. The entry point is one constructor:

\begin{codeblock}[]
from circuitkit import Pipeline

# A jailbreak-detection CSV with columns:
#   system_jailbreak, benign_req, harmful_req, clean_ans, corrupt_ans
CLEAN_PROMPT = ("System: {system_jailbreak}\n"
                "User: {benign_req}\n"
                "Please answer with only 'Yes' or 'No'.\nAssistant:")
CORRUPT_PROMPT = ("System: {system_jailbreak}\n"
                  "User: {harmful_req}\n"
                  "Please answer with only 'Yes' or 'No'.\nAssistant:")

pipe = Pipeline.from_custom_data(
    "Qwen/Qwen2.5-1.5B-Instruct",
    data_path="jailbreak_binary.csv",
    clean_prompt=CLEAN_PROMPT,   corrupt_prompt=CORRUPT_PROMPT,
    clean_answer="{clean_ans}",  corrupt_answer="{corrupt_ans}",
    task_name="jailbreak_paired",
)
pipe.discover(algorithm="eap-ig", level="node", sparsity=0.3, n_examples=256)
\end{codeblock}

\medskip
\noindent The same dataset serves the clean-only algorithms simply by omitting the corrupt templates. \emph{CircuitKIT} then routes the task to IBCircuit or CD-T, which need only the clean prompt, so a user with no counterfactual data is not locked out of discovery:

\begin{codeblock}[]
pipe = Pipeline.from_custom_data(
    "Qwen/Qwen2.5-1.5B-Instruct",
    data_path="jailbreak_binary.csv",
    clean_prompt=CLEAN_PROMPT, clean_answer="{clean_ans}",
    # no corrupt_prompt / corrupt_answer  ->  clean-only routing
    task_name="jailbreak_clean_only",
)
pipe.discover(algorithm="ibcircuit", level="node", sparsity=0.3,
              n_examples=256, num_epochs=1000, learning_rate=0.05)
\end{codeblock}

\medskip
\noindent This paired-versus-clean-only split is the same contrast that Section~\ref{subsec:e4} studies end to end. Under the hood, the template normalizer performs token-length alignment because EAP and EAP-IG compute element-wise differences between clean and corrupt activations at each sequence position, which requires the two sequences to share a token count and target position. Three alignment strategies are available: \texttt{filter} (drop misaligned pairs, the default when a tokenizer is present), \texttt{pad\_question} (pad neutral tokens into the corrupt prompt's question region up to a boundary marker), and \texttt{none}. Filtering a misaligned pair rather than padding or truncating it prevents spurious attribution signal from entering the scores; the path depends on this guarantee, and it is not a convenience feature.

\paragraph{Automatic ingestion from HuggingFace.} When the dataset already lives on the HuggingFace Hub, the templates can be skipped: \texttt{auto\_task\_from\_hf} loads a dataset by name and infers its schema, mapping columns to prompt and answer roles without a hand-written template. Detection is organized around a \texttt{DatasetShape} taxonomy that names the native shape of a raw dataset before normalization (question-answer, multiple-choice, classification, pairwise, forget-retain, refusal, and clean-only among them), which fixes how each shape becomes a discovery task. Before any discovery compute is spent, an ingested dataset passes through a worthiness pre-flight gate. Eight core checks grade it: token-count alignment between the clean and corrupt halves, single-token answer determinism, whether the model already answers correctly often enough, logit-difference signal, class balance, pair uniqueness, and corruption length-contract consistency, with further shape-specific checks. The gate returns an overall \textsc{green}, \textsc{yellow}, or \textsc{red} verdict with per-check fixes: \textsc{green} passes the configured compatibility checks, \textsc{yellow} warns that discovery may be noisy, and \textsc{red} marks a dataset unsuitable for discovery before the cost is paid. This makes ``is this dataset worth discovering on?'' a checked step rather than a post-hoc surprise.

The gate is exposed as a CLI command that grades a HuggingFace dataset before committing discovery compute and reports which algorithms are compatible with the detected data conditions.
\begin{codeblock}[listing options={language=bash}]
# Grade a HuggingFace dataset before spending any discovery compute
circuitkit data check cais/mmlu \
    --hf-subset high_school_world_history \
    --model EleutherAI/pythia-1.4b --max-records 128
# cross-model: --model Qwen/Qwen2.5-1.5B-Instruct   (or meta-llama/Llama-3.2-1B)
# prints: detected shape, a GREEN / YELLOW / RED verdict with per-check fixes,
# and artifact_safe_for: [acdc, cd-t, eap, eap-ifr, eap-ig, ibcircuit]
\end{codeblock}

When the dataset already lives on the Hub, one call infers its schema and returns a task that registers and discovers exactly like a template-defined one, with no hand-written prompt or answer template.
\begin{codeblock}[]
from circuitkit import Pipeline, register_task
from circuitkit.tasks import auto_task_from_hf

# Load a HuggingFace dataset and infer its schema -- no templates
task = auto_task_from_hf("glue", subset="sst2", split="validation")
register_task(task)

pipe = Pipeline("meta-llama/Llama-3.2-1B", task=task.name)
pipe.discover(algorithm="eap-ig", level="node", sparsity=0.3)
\end{codeblock}

\subsection{Synthesizing Corrupt Counterparts}
\label{subsec:corruption}

When the user has clean data but no natural counterfactual and still wants to run a contrastive algorithm, the corruption package can synthesize the corrupt half. It ships ten strategy classes conforming to a common protocol; five are wired into the declarative \texttt{corruption.strategy} field, and the rest are reachable by constructing a \texttt{CorruptionPipeline} directly. Three of the wired strategies are \emph{answer-changing}, intended to produce the discovery-time counterfactual, and two are \emph{answer-preserving}, intended for the Pillar~4 robustness test (Section~\ref{sec:evaluation}), where a circuit's causal effect should survive a superficial perturbation that leaves the answer intact.

\begin{itemize}
  \item \textsc{Entity-swap} (answer-changing): a detected named entity (via spaCy NER over \textsc{Person}, \textsc{GPE}, and related types) is replaced with a same-type entity, changing the target answer while typically preserving token count. Suited to QA and factual-recall tasks.
  \item \textsc{Token-swap} (answer-changing): a part-of-speech-aware substitution of a content token with another single-token word of the same tag, validated to remain one token under the model's tokenizer. Suited to classification and sentiment tasks.
  \item \textsc{Role-swap} (answer-changing): an SVO-to-OVS reordering via dependency parsing that reverses the agent-patient relationship. Suited to tasks where argument structure determines the answer.
  \item \textsc{Paraphrase} (answer-preserving): a surface rewording that pins the answer span while altering syntax, for robustness evaluation.
  \item \textsc{Distractor} (answer-preserving): injects an irrelevant-but-plausible sentence while leaving the answer unchanged, again for robustness evaluation.
\end{itemize}

\begin{figure}[!htbp]
    \centering
    \includegraphics[width=\linewidth]{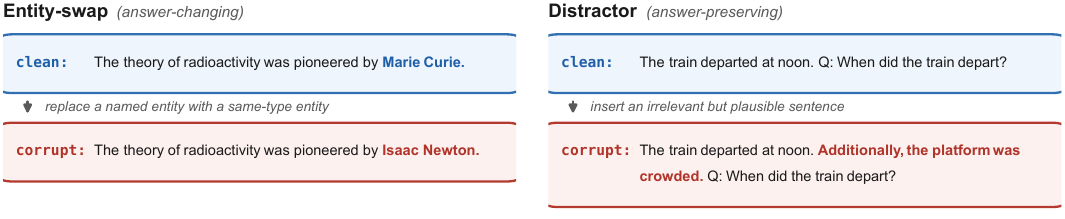}
    \caption{One strategy from each corruption mode.
    \textsc{Entity-swap} (left) edits the span the answer depends on, so the
    pair works as a discovery-time counterfactual; \textsc{Distractor} (right)
    leaves the answer intact and belongs in the Pillar~4 robustness test. Note
    the distractor also changes token count, so the alignment step of
    Section~\ref{subsec:custom-data} filters such pairs before attribution.
    Prompts are illustrative; both strategies build their edits at runtime from
    NER over the user's own data.}
    \label{fig:corruption-examples}
\end{figure}

\noindent The remaining five classes (\textsc{color-swap}, \textsc{voice-swap}, \textsc{negation}, \textsc{distractor-variation}, and \textsc{position-shift}) extend coverage to further perturbation types and are reachable programmatically. Figure~\ref{fig:corruption-examples} contrasts an answer-changing and an answer-preserving strategy. Because an answer-preserving pair carries no discriminating signal for attribution patching, using \textsc{Paraphrase} or \textsc{Distractor} where a discovery-time counterfactual is required would silently yield a clean-equals-corrupt pair; the loader detects this and logs a warning (or raises, on the stricter normalized path) rather than biasing the scores downstream.

Synthesized pairs are not trusted blindly. Beyond the degenerate-pair check on the basic path, a caller can construct a \texttt{CorruptionPipeline} with an explicit validator chain (tokenization and length-budget checks, label-consistency checks, and an embedding-based semantic-shift check) and read its per-dataset corruption-effectiveness report (mean behavioral impact, label consistency, semantic shift) to quantify pair quality before spending discovery compute.

\subsection{Programmatic Tasks}
\label{subsec:programmatic-tasks}

For multi-turn harnesses or generation-time metrics that neither the registry nor the template path expresses, a user implements a small module conforming to the \texttt{TaskSpec} protocol, supplying a dataloader builder and a metric function. The WMDP unlearning benchmark~\cite{li2024wmdp} began as such a module (roughly fifty lines) and was later promoted into the built-in registry, an instance of the intended lifecycle from user-contributed \texttt{TaskSpec} to first-class task.

\section{Faithfulness Evaluation}
\label{sec:evaluation}

A discovered circuit is only useful if it faithfully explains the behavior it was found for. \emph{CircuitKIT} evaluates a circuit along six configurable pillars, motivated by the finding that single faithfulness metrics are not robust across ablation methods and can even reverse method rankings~\cite{miller2024faithfulness}. Each pillar captures a distinct failure mode; together they give a multi-faceted assessment that no single score supplies. All six run from one entry point, and any subset is configuration-selectable:

\begin{codeblock}[]
report = pipe.evaluate(
    pillars=["patching", "ablation", "stability",
             "robustness", "baselines", "generalization"],
    n_stability_runs=3,
)
pipe.summary()   # formatted per-pillar table
\end{codeblock}

\paragraph{Notation.} Write $m$ for the task metric (logit difference by default), $C$ for the discovered circuit, and $y_{\text{clean}}, y_{\text{corrupt}}$ for $m$ on the \emph{full} model under the clean and corrupt inputs; $y_{\text{circuit}}$ is $m$ evaluated with only $C$ active and its complement intervened on. Pillars~1 and~2, and the transfer ratio of Pillar~6, share one normalized faithfulness score,
\begin{equation}
    F(C) \;=\; \frac{y_{\text{circuit}} - y_{\text{corrupt}}}{y_{\text{clean}} - y_{\text{corrupt}}},
    \label{eq:faithfulness}
\end{equation}
equal to $1$ when $C$ recovers full-model behavior and $0$ at the corrupt baseline; the two variants below differ only in how $y_{\text{circuit}}$ is produced. $F$ is reported \emph{signed} (an optional clip to $[0,1]$ is exposed) and \emph{invalid} rather than numeric when the denominator is degenerate ($|y_{\text{clean}} - y_{\text{corrupt}}| < \varepsilon$) or inverted, since that reflects the metric's direction, not the circuit~\cite{miller2024faithfulness}.

\paragraph{Pillar 1: Causal Patching.} Whether the circuit \emph{causally explains} the behavior. A clean forward pass is run while every edge \emph{outside} the circuit is patched toward its corrupted-run activation (noising the complement rather than denoising the circuit), and the score is $F$ (Eq.~\ref{eq:faithfulness}) with $y_{\text{circuit}}$ produced by this \emph{soft} intervention~\cite{mueller2025mib}. Higher is more sufficient.

\paragraph{Pillar 2: Ablation Faithfulness.} Whether the circuit is sufficient \emph{in isolation}. The circuit is kept and every out-of-circuit component is ablated (zero, mean, or mean-positional replacement); the score is $F$ (Eq.~\ref{eq:faithfulness}) with $y_{\text{circuit}}$ produced by this \emph{hard} ablation. Pillar 1's counterfactual is \emph{soft} (the complement is pushed toward its corrupted-run activations); ablation's is \emph{hard} (outright removal), and because absolute ablation scores depend on the replacement distribution~\cite{miller2024faithfulness}, the library exposes multiple ablation methods and reports each instead of committing to one. Ratios are reported \emph{signed}: a value below zero means the isolated circuit was driven beneath the corrupt-run baseline, a behavioral inversion rather than a measurement error, and Sections~\ref{subsec:e3} and~\ref{subsec:e4} discuss such cells on real data.

\paragraph{Pillar 3: Stability.} The \emph{reproducibility} of discovery across independent data resamples. Given circuits $C_1, C_2$ found from different random samples of the same task with the same method, with score vectors $\mathbf{s}_1, \mathbf{s}_2$,
\begin{equation}
    J(C_1, C_2) = \frac{|C_1 \cap C_2|}{|C_1 \cup C_2|}, \qquad
    \rho(C_1, C_2) = \operatorname{Spearman}(\mathbf{s}_1, \mathbf{s}_2),
    \label{eq:jaccard}
\end{equation}
where Jaccard $J$ is set-level overlap and Spearman $\rho$ is threshold-free rank agreement. High stability means the mechanism reflects persistent model structure, not sampling noise; stability is more than hygiene, since a circuit's stability has itself been argued to characterize how a model generalizes~\cite{anon2025stability}.

\paragraph{Pillar 4: Robustness.} Whether the circuit's causal-patching score holds across the corruption-strategy family of Section~\ref{subsec:corruption}: the answer-preserving strategies (paraphrase, distractor), where the score should be unchanged, and the answer-changing ones (entity-swap, token-swap, role-swap), which test the circuit against a different valid counterfactual. A robust circuit's explanatory power survives the whole family, not one specific corruption. Per variant $c$ in the family $\mathcal{K}$, the pillar reports the ratio and relative drop of the circuit's patching score,
\begin{equation}
    r_c(C) = \frac{y_{\text{circuit}}^{(c)}}{y_{\text{circuit}}^{(0)}}, \qquad
    \delta_c(C) = 1 - r_c(C), \qquad c \in \mathcal{K},
    \label{eq:robustness}
\end{equation}
where $y_{\text{circuit}}^{(0)}$ is the score on the original data; a robust circuit keeps $r_c \approx 1$, and each $r_c$ is reported per variant rather than reduced to a single scalar.

\paragraph{Pillar 5: Baseline Comparison.} Whether the circuit beats structurally comparable non-causal baselines (random subgraphs and weight-magnitude selections of equal size), that is, whether its structure is meaningful rather than incidental sparsity. For each size-matched baseline $b$ (a random subgraph averaged over $n$ draws, or a weight-magnitude selection) the pillar reports an improvement ratio and a random-baseline $z$-score,
\begin{equation}
    I_b(C) = \frac{y_{\text{circuit}}}{y^{(b)}_{\text{baseline}}}\;\;(y^{(b)}_{\text{baseline}} > 0),
    \qquad
    z = \frac{y_{\text{circuit}} - \bar y_{\text{rand}}}{\sigma_{\text{rand}}},
    \label{eq:baselines}
\end{equation}
where $I_b$ is reported as a status flag ($\dagger$) rather than a number when the baseline metric is non-positive, and $(\bar y_{\text{rand}}, \sigma_{\text{rand}})$ are the mean and standard deviation over the random draws.

\paragraph{Pillar 6: Generalization.} Whether a circuit discovered on one task retains predictive structure on a related task without rediscovery. This pillar requires an \emph{adjacent} dataset that exercises the same behavior on different specifics: the library pairs IOI with \texttt{double\_io} (indirect-object identification over two-sentence contexts) and pairs one WMDP domain with another (WMDP-Cybersecurity transferring to WMDP-Biology). Writing $F_{\text{source}}, F_{\text{target}}$ for the normalized faithfulness (Eq.~\ref{eq:faithfulness}) on the two tasks, the transfer score is
\begin{equation}
    T(C) = \frac{F_{\text{target}}(C)}{F_{\text{source}}(C)}, \qquad \text{invalid if } F_{\text{source}}(C) < \varepsilon,
    \label{eq:generalization}
\end{equation}
reported as invalid rather than a spurious number when the source faithfulness is degenerate or non-positive; the raw-metric variant $y_{\text{target}}/y_{\text{source}}$ is the default, and the normalized form is recommended for signed metrics.

The cross-task transfer matrix discovers a circuit on each source task and scores it on every target task, exposing Pillar 6 generalization directly as a NumPy matrix plus a best/worst-transfer summary.
\begin{codeblock}[]
from circuitkit import load_model
from circuitkit.evaluation.transfer import TransferMatrix

model = load_model("google/gemma-2-2b")  # or "meta-llama/Llama-3.2-1B" / "Qwen/Qwen2.5-1.5B-Instruct"
matrix = TransferMatrix(task_names=["ioi", "sva", "greater_than"])
scores = matrix.build(model, discovery_cfg_template={
    "model": {"name": "google/gemma-2-2b", "precision": "float32"},
    "discovery": {"algorithm": "eap-ig", "data_params": {"num_examples": 16}},
    "pruning": {"target_sparsity": 0.3}}, device="auto")
analysis = matrix.analyze()  # {"best_transfer": (src, tgt, score), "worst_transfer": ...}
# CLI: circuitkit transfer-matrix -m google/gemma-2-2b -t ioi,sva,greater_than
\end{codeblock}

\paragraph{Pillar 7 (optional): Intervention Reliability.} Beyond the six core pillars, the framework exposes a seventh that measures how consistently a circuit-guided \emph{intervention} produces its intended effect across re-discoveries. It combines three sub-scores: seed consistency (Spearman $\rho$ of per-component scores across re-runs), effect magnitude (the normalized intervention effect relative to baseline), and effect variance (one minus the coefficient of variation of that effect), aggregated into a reliability index in $[0,1]$. The three map to $R_1 = (\bar\rho+1)/2$, $R_2$ (the mean effect magnitude relative to baseline, mapped to $[0,1]$), and $R_3 = \operatorname{clip}_{[0,1]}\!\big(1 - \operatorname{SD}(\Delta)/|\overline{\Delta}|\big)$, combined by their harmonic mean,
\begin{equation}
    \mathrm{RI}(C) = \frac{3}{R_1^{-1} + R_2^{-1} + R_3^{-1}} \in [0,1].
    \label{eq:reliability}
\end{equation}
It is optional, and not one of the six core pillars, because unlike Pillars 1 through 6 it requires re-running discovery rather than scoring a single circuit. The cross-seed dispersion statistic it reports,
\begin{equation}
    \text{SD}_{\text{rel}} = \operatorname{SD}\!\left(\Delta_{\text{metric}}\right)_{n},
    \label{eq:sdrel}
\end{equation}
the sample standard deviation of the downstream metric across $n$ independent re-discoveries, is the designated statistic for quantifying discovery-induced variance in an intervention's effect; the single-seed intervention studies of Section~\ref{sec:experiments} leave measuring it at scale to future work.

\section{Intervention Modules}
\label{sec:applications}

Once a circuit is discovered and evaluated, \emph{CircuitKIT}'s intervention modules use it to modify the model. Every module consumes the same \texttt{CircuitScores} artifact produced by any backend, so switching the discovery algorithm behind an intervention is selected through a common configuration interface. These modules are what make circuit analysis \emph{actionable} rather than merely descriptive, and they do so by feeding circuit-derived importance into established tooling rather than reimplementing it: structural pruning runs through LLM-Pruner~\cite{ma2023llmpruner}, mixed-precision quantization through \texttt{quanto} and \texttt{llmcompressor}, selective fine-tuning through PEFT/LoRA~\cite{hu2022lora}, and knowledge editing through ROME and MEMIT~\cite{meng2022rome, meng2023memit}, with every resulting checkpoint scored on standardized benchmarks through \texttt{lm-evaluation-harness}~\cite{biderman2024lessons}. Three modules are the primary intervention layer, exercised across the empirical studies of Section~\ref{sec:experiments}: structural pruning, mixed-precision quantization, and selective fine-tuning. Four further modules (knowledge editing, activation steering, circuit-restricted LoRA healing, and hallucination probing) ship in the library and are described at the end of the section.

\subsection{Structural Pruning}
\label{subsec:pruning}

Structural pruning removes the model components a circuit judges least important. The flat API's \texttt{prune()} zero-masks the lowest-scoring attention heads and MLP components in place; because tensor shapes are unchanged, masking alone does not reduce parameter count or FLOPs until the model is written out, at which point \texttt{export\_checkpoint} physically removes the masked channels and writes a reloadable HuggingFace checkpoint:

\begin{codeblock}[]
circuit = ck.load_scores("./circuit.pt")
pruned  = ck.prune(model, circuit, sparsity=0.3, scope="both")
ck.export_checkpoint(pruned, circuit, "./output/pruned_checkpoint")
\end{codeblock}

\medskip
\noindent For neuron-level pruning that composes with an established framework, the circuit's attribution scores are wrapped as a custom importance estimator and passed to LLM-Pruner's~\cite{ma2023llmpruner} \texttt{MetaPruner}, which removes whole attention heads (as consecutive groups on the key-projection root) and per-neuron MLP channels (on the gate projection). This makes circuit-derived importance a drop-in criterion alongside LLM-Pruner's native ones, magnitude ($\ell_2$ weight norm) and Taylor (first-order loss expansion), so a circuit-guided cut can be compared against them directly. The essential difference is the source of the importance signal: circuit-guided pruning ranks components by \emph{causal attribution of a specific behavior} rather than by aggregate weight statistics or generic loss gradients. Section~\ref{subsec:e5} reports this comparison.

\subsection{Mixed-Precision Quantization}
\label{subsec:quantization}

Under a fixed precision budget, some layers must be quantized more aggressively than others, and the question is which to protect. \emph{CircuitKIT} answers it with circuit importance: components are ranked (attention and MLP independently, by default), the top fraction is kept at native precision, and the rest are quantized:

\begin{codeblock}[]
plan = ck.quantize(hf_model, circuit,
                   high_fraction=0.3,   # top 30% of layers stay full precision
                   backend="quanto")    # or "llmcompressor" for true sub-4-bit
\end{codeblock}

\medskip
\noindent The default \texttt{quanto} backend assigns integer precision tiers internally; pinning an explicit bit-width uses the \texttt{llmcompressor} backend (GPTQ-calibrated, vLLM-compatible). Because the tier-selection rule (aggregate a circuit score per component, then split into a high and a low tier under the budget) is independent of which low-bit kernel fills the quantized tier, circuit-guided selection can be compared cleanly against uniform quantization and random-tier baselines at matched budget. Section~\ref{subsec:e6} reports this comparison.

\subsection{Selective Fine-Tuning}
\label{subsec:finetuning}

Selective fine-tuning updates only the parameters a circuit marks as important, freezing the rest. \texttt{selective\_finetune()} resolves the top-$k\%$ most important attention heads and MLP components to concrete weight-matrix index ranges and returns a selection rather than a trained model:

\begin{codeblock}[]
result = ck.selective_finetune(circuit, model_name="Qwen/Qwen2.5-1.5B-Instruct",
                               top_fraction=0.2, scope="both")
# result.attn: {"attn_4": {"q": [...], "k": [...], ...}, ...}
# result.mlp:  {"mlp_2": [neuron indices] or None, ...}
\end{codeblock}

\medskip
\noindent The caller wires this selection into either a static gradient mask (only the selected positions receive gradient updates; everything else is frozen) or a PEFT \texttt{LoraConfig}'s \texttt{target\_modules}. The comparison this enables is against random-budget selection and full-parameter training (or matched-budget LoRA~\cite{hu2022lora}) under an identical training recipe, isolating the effect of \emph{which} parameters are updated. Section~\ref{subsec:e7} reports the static-mask comparison.

\subsection{Additional Intervention Modules}
\label{subsec:additional-interventions}

Four further modules ship in the library and consume the same artifact through the same API patterns.

\paragraph{Knowledge editing.} \emph{CircuitKIT} wraps ROME~\cite{meng2022rome} for single-fact edits and MEMIT~\cite{meng2023memit} for batch edits, both restricted to circuit-resident MLP layers, and adds circuit-aware variants (CaKE, multi-hop MCircKE, and a gradient fine-tuning fallback). ROME and MEMIT insert a rank-one update into an MLP weight matrix, and the solution for the update direction requires the second-moment statistic $C \triangleq \lambda\,\mathbb{E}[k\,k^\top]$ of the layer's key vectors over a representative corpus (ROME solves $(C + \lambda I)\,v = k$ for the edit direction). This matrix is \emph{model- and layer-specific}: it depends on the checkpoint's own activation statistics, so it cannot be shipped as a fixed asset. \emph{CircuitKIT} provides \texttt{get\_covariance}, which estimates and caches $C$ from a text corpus the user supplies, keyed by (model, layer, sample count), so the cost is paid once per model and reused across every subsequent edit:

\begin{codeblock}[]
from circuitkit.applications.editing.rome_wrapper import RomeHandler
from circuitkit.applications.common_utils._covariance import get_covariance

# Estimate C once from a representative corpus; cached on disk per (model, layer).
C = get_covariance(model, layer=circuit_mlp_layer,
                   hook_name="mlp.hook_post",
                   texts=my_corpus, n_samples=100_000)

editor = RomeHandler(model)
editor.edit_single_fact("The capital of France is", "France", "Berlin",
                        target_layer=circuit_mlp_layer)
\end{codeblock}

\medskip
\noindent The original ROME and MEMIT papers estimate $C$ from roughly one hundred thousand Wikipedia samples; \emph{CircuitKIT} ships a small fallback corpus adequate only for a smoke test, and we recommend supplying a corpus of comparable scale for edit-quality-sensitive use. Making corpus estimation a one-call, cached step turns a model-specific prerequisite into a manageable one: an edit on a new checkpoint needs no changes beyond pointing \texttt{get\_covariance} at a corpus. For unlearning use cases such as WMDP (Section~\ref{subsec:programmatic-tasks}), the editing module also ships an \texttt{UnlearningVerifier} that checks whether a circuit-scoped edit actually removed the targeted knowledge and returns a structured \texttt{UnlearningReport}; it is a verification helper on the same knowledge-editing module, not a separate counted intervention.

If you do not want to manage the layer index or handler wiring, \texttt{CircuitKnowledgeEditor.edit\_via\_circuit} auto-selects the circuit-resident MLP layer and dispatches to ROME/MEMIT/FT behind a single call.
\begin{codeblock}[]
from circuitkit import load_model
from circuitkit.applications.editing import CircuitKnowledgeEditor

model = load_model("meta-llama/Llama-3.2-3B-Instruct")
editor = CircuitKnowledgeEditor(model)

# circuit=None -> auto-selects the middle (circuit-resident) MLP layer.
result = editor.edit_via_circuit("The capital of France is",
                                 subject="France", target="Lyon", method="rome")
print(result.success, result.confidence_before, result.confidence_after, result.target_layer)
\end{codeblock}

\paragraph{Activation steering.} \texttt{ActivationSteering} learns a steering vector at each high-scoring circuit node from source and target example pairs (the difference of their mean activations) and adds a scaled copy at inference time, following the activation-addition construction of~\cite{turner2025steering}; a negative coefficient steers away from the target direction. It attaches at \texttt{attn.hook\_result}, so it operates on the same per-head channels that discovery scores.

A typical use consumes a discovered circuit's score dict, learns the steering vectors from source and target example pairs, and applies them at inference with a single coefficient.
\begin{codeblock}[]
from circuitkit import load_model, load_scores
from circuitkit.applications.steering import ActivationSteering

model = load_model("Qwen/Qwen2.5-1.5B-Instruct")  # or "gpt2"
circuit = load_scores("results/circuit.pt")       # circuit.scores: name -> score

steering = ActivationSteering(model, circuit.scores, score_threshold=0.5)
vectors = steering.compute_steering_vector(source_examples, target_examples)
result = steering.steer("When Alice and Bob walked in,", vectors, coefficient=1.5)
logits = result["output"]  # negative coefficient steers away from the target
\end{codeblock}

\paragraph{Circuit-restricted LoRA healing.} After structural pruning degrades a model, \texttt{CircuitLoRA} trains a small LoRA adapter on the pruned model but grants adapter parameters only to modules whose circuit score exceeds a threshold, concentrating recovery capacity on the components discovery judged most causally important. Because it composes with an already-discovered pruning circuit, it recovers a fraction of the pruning gap at no additional discovery cost.

\paragraph{Hallucination probing.} \texttt{HallucinationDetector} trains a per-layer linear probe on circuit activations from labeled factual and hallucinated examples, then flags probable hallucinations in new text, giving an off-circuit diagnostic use of a discovered circuit.

\section{Visualization, Reporting, and Benchmarking}
\label{sec:visualization}

The final stage turns a discovered circuit into something a human can inspect and a number a paper can cite. All of it runs off the same \texttt{CircuitScores} artifact.

\subsection{Visualization}
\label{subsec:viz}

\texttt{ck.visualize\_circuit} renders a circuit in one of three modes. The \texttt{graph} mode draws the discovered subgraph as an annotated node-and-edge diagram, nodes sized and colored by importance and hoverable for detail, exported as a self-contained zoom/pan/threshold-slider HTML page. The \texttt{comparison} mode places two or more circuits side by side, from different seeds, algorithms, or tasks, and renders stability heatmaps, score-correlation matrices, and distribution plots that make method or seed agreement visible at a glance. The \texttt{dashboard} mode launches an interactive Streamlit application over the same renderers, and a separate Jupyter widget suite exposes them inline for exploratory work.

\begin{codeblock}[]
# Single-circuit interactive graph
ck.visualize_circuit(circuit, mode="graph", output="circuit.html")

# Side-by-side comparison of two circuits (heatmap + correlation)
ck.visualize_circuit(circuit_a, mode="comparison", second_circuit=circuit_b, output="comp.html")
\end{codeblock}

\begin{figure}[!htbp]
    \centering
    \includegraphics[width=0.6\linewidth]{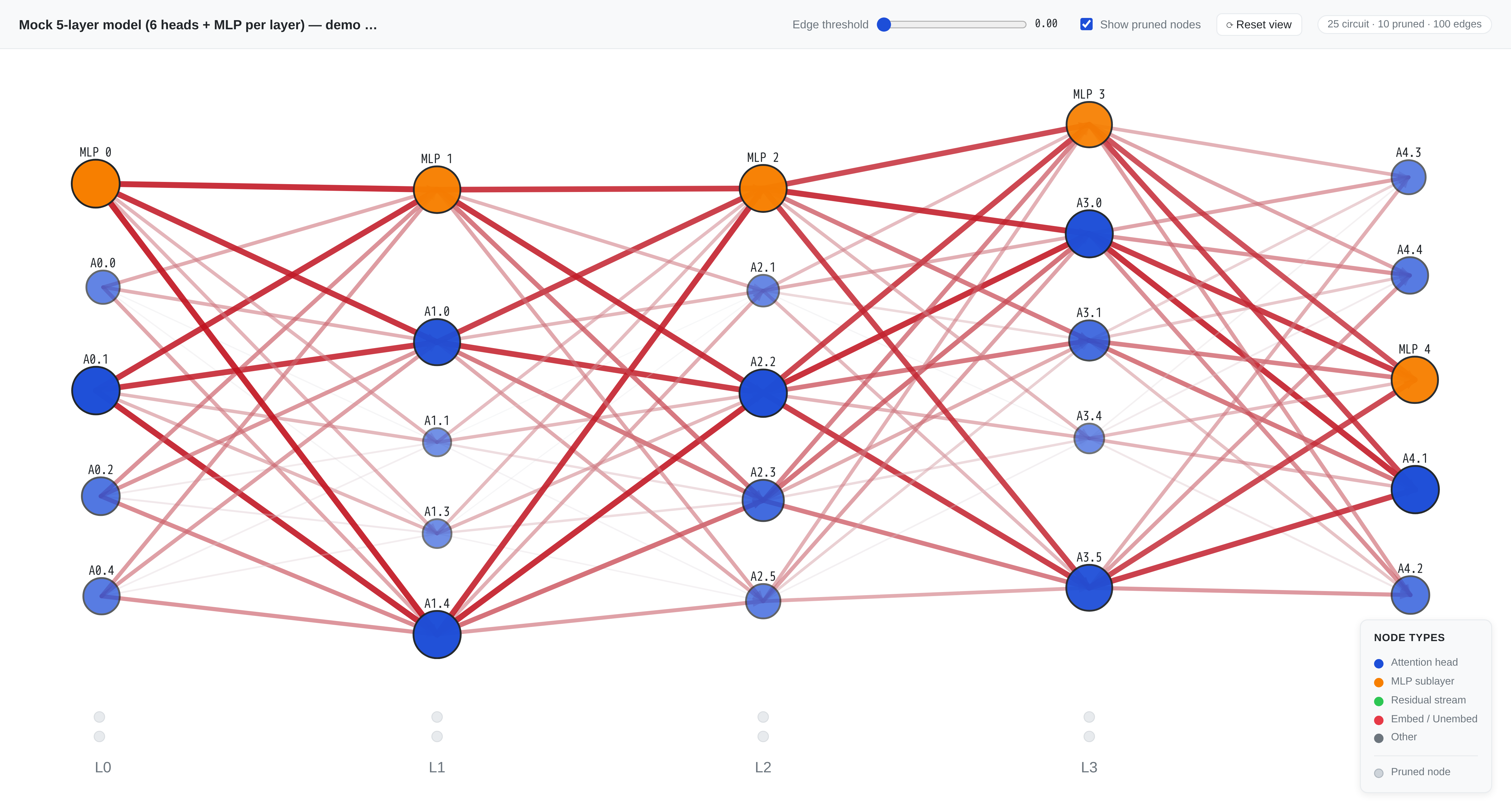}
    \caption{\texttt{ck.visualize\_circuit(mode="graph")}, real output; a small five-layer model (${\sim}30\%$ pruned) is shown for print legibility. Node color is component type, node size is attribution score, and edge opacity is edge score. The live artifact is a self-contained HTML page with zoom, tooltips, and the edge-threshold and pruned-node controls visible along its top edge.}
    \label{fig:viz-graph}
\end{figure}

\medskip
\noindent Figure~\ref{fig:viz-graph} shows the discovered-circuit graph. Beyond it, the library ships the comparison-mode stability heatmap and score-correlation matrix, activation- and feature-saliency maps (token-and-position heatmaps across layers), and an interactive circuit editor for adding, removing, or toggling nodes and edges; we mention these rather than depict them, since the graph and reporting views are the ones the studies below rely on.

\FloatBarrier

\subsection{Reporting}
\label{subsec:reporting}

\begin{figure}[!htbp]
    \centering
    \includegraphics[width=0.33\linewidth]{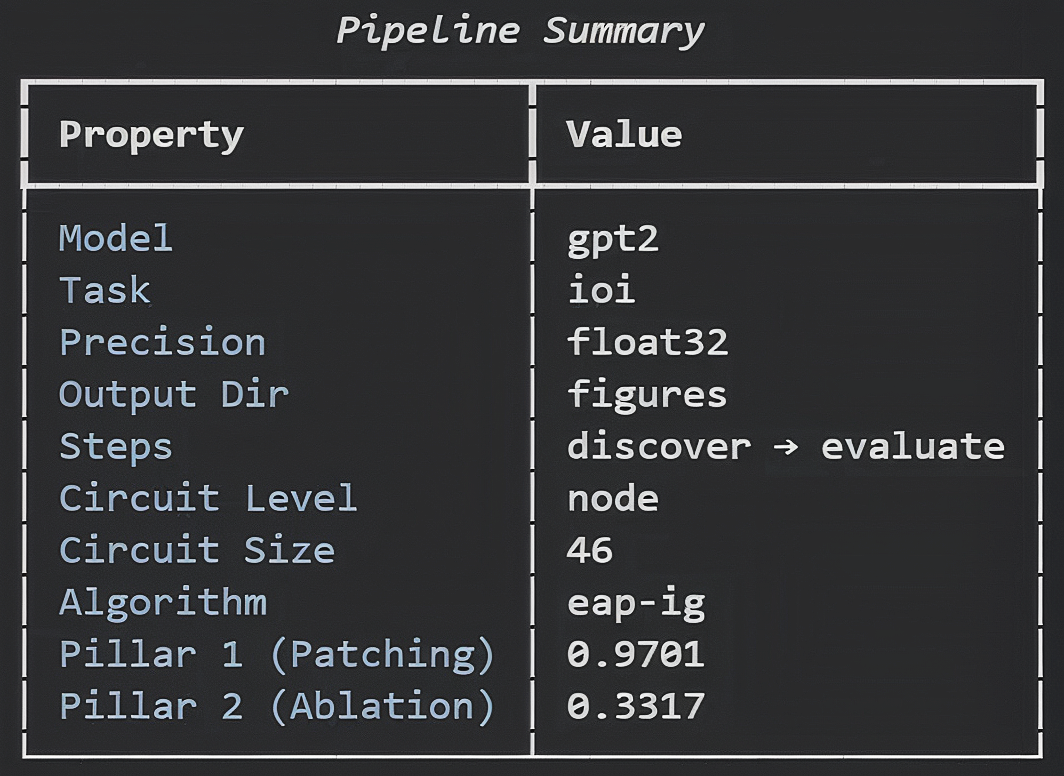}
    \caption{\texttt{pipe.summary()} as rendered in the terminal, captured from
    a live GPT-2\,/\,IOI \texttt{eap-ig} run at node level. The same
    \texttt{CircuitScores} record that drives the interactive graph of
    Figure~\ref{fig:viz-graph} renders here, and serializes to JSON unchanged.}
    \label{fig:pipe-summary}
\end{figure}

Evaluation runs produce human-readable reports. \texttt{pipe.summary()} prints the pipeline state and pillar scores for the current circuit as a formatted terminal table (Figure~\ref{fig:pipe-summary}); the underlying \texttt{FaithfulnessReport} serializes to JSON so a run's full pillar breakdown can be logged, diffed across configurations, or embedded in a downstream report. Because the report is keyed by the same task, model, and algorithm fields the artifact carries, a batch of runs aggregates into a comparison grid without any bookkeeping on the caller's side. The four calls that produce Figure~\ref{fig:pipe-summary} are the run itself:

\begin{codeblock}[]
pipe = Pipeline("gpt2", task="ioi")
pipe.discover(algorithm="eap-ig", level="node")
pipe.evaluate(pillars=["patching", "ablation"])
pipe.summary()   # prints the formatted panel below
\end{codeblock}

\FloatBarrier

\section{Extensibility}
\label{sec:extensibility}

\emph{CircuitKIT} is built to be extended along every axis on which mechanistic interpretability moves, and it uses one pattern for all of them: a typed registry with a decorator entry point. New capability is added by registering, not by editing the core.

\paragraph{Selector registry.} At the center of the intervention layer sits a registry of \emph{selectors}: importance-scoring functions with the uniform signature \texttt{(model, task, config) $\to$ Dict[str, float]}. Fourteen ship built in, spanning circuit-discovery attribution (\texttt{eap}, \texttt{eap-ig}, \texttt{eap-gp}, \texttt{ibcircuit}, \texttt{cdt}, \texttt{relp}), pruning and quantization baselines (\texttt{magnitude}, \texttt{taylor}, \texttt{wanda}, \texttt{gptq}, and calibration-based variants), and a random control. Because a selector is just a scoring function, any application module can use any selector interchangeably as its importance criterion, which is what makes the circuit-versus-baseline comparisons in Section~\ref{sec:experiments} a matter of swapping one name. Whether an attribution selector genuinely ranks the components that causally drive a behavior, rather than merely the rank-stable ones, is itself an active question~\cite{eswar2026refusal}. A new selector is one decorator:

\begin{codeblock}[]
from circuitkit.selection import register

@register("my_selector")
def my_selector(model, task_name, config) -> dict:
    # return {component_name: importance_score, ...}
    ...
\end{codeblock}

\paragraph{Other registries.} The same pattern governs the discovery backends (the algorithm registry that is the single source of truth for names and stability tiers, Section~\ref{sec:discovery}), the corruption strategies (all ten conform to one \texttt{CorruptionStrategy} protocol, Section~\ref{subsec:corruption}), and the model families for interventions (the architecture registry, Section~\ref{subsec:coverage}). This gives a clear governance model: stable-tier methods are maintained by the core team with backward-compatibility guarantees, while research-tier and community contributions register alongside them without touching the library core, and are promoted as validation evidence accrues.

A new corruption strategy is likewise one decorator; it declares a \texttt{LengthContract} so that gradient-based discovery, which needs token-aligned clean and corrupt passes, can verify the pair shapes before running.
\begin{codeblock}[]
from circuitkit.data.corruption import (
    CorruptionStrategy, CorruptionResult, register_strategy)
from circuitkit.data.corruption.base import LengthContract

@register_strategy("synonym_swap")
class SynonymSwap(CorruptionStrategy):
    length_contract = LengthContract.PRESERVE  # token-aligned pairs for EAP
    def corrupt(self, record, **kwargs) -> CorruptionResult:
        prompt = record.clean_prompt.replace("Alice", "Anna", 1)
        return CorruptionResult(corrupt_prompt=prompt,
                                corrupt_answer=record.clean_answer)
\end{codeblock}

\subsection{Benchmarking}
\label{subsec:benchmarking}

The loop closes at measured downstream quality. \texttt{ck.benchmark} runs an exported checkpoint through \texttt{lm-evaluation-harness}~\cite{biderman2024lessons}, so a pruned or quantized model is scored on the same standardized benchmarks as any other HuggingFace checkpoint~\cite{wolf-etal-2020-transformers}. Every downstream accuracy in Section~\ref{sec:experiments} is produced through this integration, which is what lets the intervention studies report standardized benchmark accuracy instead of proxy metrics:

\begin{codeblock}[]
ck.export_checkpoint(pruned, circuit, "./output/pruned_checkpoint")
scores = ck.benchmark("./output/pruned_checkpoint", tasks=["mmlu"], num_fewshot=5)
\end{codeblock}

\medskip
\noindent Because the exported artifact is a standard HuggingFace checkpoint, the same \texttt{ck.benchmark} call can route through \texttt{lm-evaluation-harness}'s native vLLM backend, scoring a pruned or quantized checkpoint at higher throughput on larger models with no change to the pipeline.

\section{Using \emph{CircuitKIT}}
\label{sec:usage}

\emph{CircuitKIT} exposes the full discover, evaluate, and intervene pipeline through three interfaces, all producing and consuming the same \texttt{CircuitScores} artifact. Because the artifacts are interchangeable, a circuit discovered interactively in a notebook can be evaluated from the CLI and consumed by a pruning script without any conversion. Figure~\ref{fig:e2e} drives the same pipeline four ways, end to end: switching the task, the data source, the intervention, or the interface changes only a field, not the program.

\begin{figure*}[htbp]
    \centering
    \begin{minipage}[t]{0.48\textwidth}
    \textbf{\footnotesize (a) Discover and evaluate a built-in task}
\begin{lstlisting}[language=Python,basicstyle=\scriptsize\ttfamily,aboveskip=2pt,frame=single,rulecolor=\color{black!25},framesep=5pt,backgroundcolor=\color{lstbg},xleftmargin=3pt,xrightmargin=3pt]
from circuitkit import Pipeline
pipe = Pipeline("meta-llama/Llama-3.2-1B",task="ioi")
pipe.discover(algorithm="eap-ig", level="node")
report = pipe.evaluate()   # six-pillar panel
pipe.summary()             # per-pillar table
\end{lstlisting}
    \end{minipage}\hfill
    \begin{minipage}[t]{0.48\textwidth}
    \textbf{\footnotesize (b) Bring a raw CSV, no contrastive pairs}
\begin{lstlisting}[language=Python,basicstyle=\scriptsize\ttfamily,aboveskip=2pt,frame=single,rulecolor=\color{black!25},framesep=5pt,backgroundcolor=\color{lstbg},xleftmargin=3pt,xrightmargin=3pt]
from circuitkit import Pipeline
pipe = Pipeline.from_custom_data("Qwen/Qwen3-1.7B",
    data_path="refusals.csv", clean_prompt="{question}",
    clean_answer="{label}")   # clean-only route
pipe.discover(algorithm="ibcircuit", level="node")

\end{lstlisting}
    \end{minipage}

    \vspace{0.4em}

    \begin{minipage}[t]{0.48\textwidth}
    \textbf{\footnotesize (c) Intervene, export, and benchmark}
\begin{lstlisting}[language=Python,basicstyle=\scriptsize\ttfamily,aboveskip=2pt,frame=single,rulecolor=\color{black!25},framesep=5pt,backgroundcolor=\color{lstbg},xleftmargin=3pt,xrightmargin=3pt]
import circuitkit as ck
pipe = ck.Pipeline("meta-llama/Llama-3.2-3B",task="boolq")
pipe.discover(algorithm="eap-ig", level="neuron")
pipe.prune(sparsity=0.3)
pipe.export("./out/pruned")   # reloadable HF ckpt
ck.benchmark("./out/pruned", tasks=["boolq"])
\end{lstlisting}
    \end{minipage}\hfill
    \begin{minipage}[t]{0.48\textwidth}
    \textbf{\footnotesize (d) Reproduce any run from the CLI / YAML}
\begin{lstlisting}[language=bash,basicstyle=\scriptsize\ttfamily,aboveskip=2pt,frame=single,rulecolor=\color{black!25},framesep=5pt,backgroundcolor=\color{lstbg},xleftmargin=3pt,xrightmargin=3pt]
# pipeline.yaml names model, task, discovery,
# evaluate, applications, export, benchmark
$ circuitkit run pipeline.yaml     # no Python
$ circuitkit inspect circuit.pt    # scores + state
$ circuitkit discover -m meta-llama/Llama-3.2-3B-Instruct \
    -a eap-ig -t boolq -s 0.3 -o circuit.pt



\end{lstlisting}
    \end{minipage}
    \caption{\emph{CircuitKIT} end to end. Every panel drives the same
    discover, evaluate, and intervene pipeline over one interchangeable
    \texttt{CircuitScores} artifact: (a) discover and evaluate a built-in task;
    (b) take a raw CSV with only clean examples, which routes to the clean-only
    algorithms; (c) intervene, export a reloadable checkpoint, and score it
    through \texttt{lm-evaluation-harness}; (d) run or reproduce the same
    pipeline from the terminal or a version-controlled YAML file. Switching the
    task, data source, intervention, or interface changes only a field, not the
    program.}
    \label{fig:e2e}
\end{figure*}

\subsection{Installation}
\label{subsec:installation}

This paper describes \emph{CircuitKIT}~v1.0.0, the first stable (production) release. The library installs from source with \texttt{pip} and requires Python~3.10 or later:

\begin{codeblock}[listing options={language=bash}]
git clone https://github.com/Lexsi-Labs/CircuitKIT.git
cd CircuitKIT
pip install -e .                    # core library
pip install -e ".[benchmarks]"      # + lm-evaluation-harness integration
pip install -e ".[quantization]"    # + optimum-quanto for circuit-guided quantization (E6)
pip install -e ".[cdt]"             # + captum/lime/shap for the CD-T discovery backend (E1)
pip install -e ".[dev]"             # test/lint tooling for contributors
\end{codeblock}

\medskip
\noindent The GPT-2/IOI quickstart below runs in a few minutes, so the framework can be tried before scaling to a multi-billion-parameter model.

\subsection{The Stateful Pipeline}
\label{subsec:pipeline-iface}

The \texttt{Pipeline} class is the recommended entry point for multi-step workflows. It loads the model lazily on the first call that needs it, carries the discovered circuit and evaluation report across method calls, and delegates to the underlying API functions. A complete discover-evaluate-prune-export-report run reads top to bottom:

\begin{codeblock}[]
from circuitkit import Pipeline
pipe = Pipeline("microsoft/phi-2", task="hypernymy")
pipe.discover(algorithm="eap-ig", n_examples=128, sparsity=0.3)
pipe.evaluate(pillars=["patching", "ablation"])
pipe.prune(sparsity=0.3)
pipe.export("./output/checkpoint")
pipe.summary()
\end{codeblock}

Because the discovered circuit is a serialized artifact, a saved run can be reloaded and reused across many intervention settings without paying for discovery again.
\begin{codeblock}[]
from circuitkit import Pipeline
# Reload a saved circuit -- no re-discovery. Cross-model: model_name="Qwen/Qwen2.5-1.5B-Instruct"
pipe = Pipeline.from_artifact("./circuit.pt", model_name="microsoft/phi-2", task="hypernymy")
for sparsity in (0.1, 0.3, 0.5):        # sweep sparsities on one artifact
    pipe.prune(sparsity=sparsity, scope="both")
    pipe.export(f"./output/prune_{sparsity}")
\end{codeblock}

\subsection{The Flat One-Shot API}
\label{subsec:flat-iface}

The same workflow is available as stateless functions in the \texttt{ck.*} namespace, for scripting and single-call use. Each function takes the artifact explicitly, so stages can be run in separate processes or interleaved with other code:

\begin{codeblock}[]
import circuitkit as ck

model   = ck.load_model("google/gemma-2-2b")
circuit = ck.discover(model, task="ioi", algorithm="eap-ig",
                      sparsity=0.3, output_path="./circuit.pt")
report  = ck.faithfulness(model, circuit, task="ioi",
                          pillars=["patching", "ablation"])
pruned  = ck.prune(model, circuit, sparsity=0.3)
ck.export_checkpoint(pruned, circuit, "./output/checkpoint")
\end{codeblock}

\subsection{The CLI and YAML}
\label{subsec:cli-iface}

A Click-based CLI drives the same pipeline from the terminal, with fifteen top-level commands plus grouped \texttt{data} and \texttt{debug} sub-commands. Single steps are one line each:

\begin{codeblock}[listing options={language=bash}]
circuitkit discover -m EleutherAI/pythia-1.4b -a eap-ig -t ioi -s 0.3 -o circuit.pt
circuitkit inspect circuit.pt
circuitkit run pipeline.yaml
\end{codeblock}

Beyond discovery and evaluation, the same CLI exposes the intervention and custom-data commands directly, so cross-task transfer, activation steering, circuit-restricted LoRA healing, and template-driven dataset construction each run as a single terminal invocation.
\begin{codeblock}[listing options={language=bash}]
# gpt2 for the light paths (swap in any HF family); Qwen for registry interventions
circuitkit transfer-matrix -m gpt2 -t ioi,sva,greater_than
circuitkit steer -m Qwen/Qwen2.5-1.5B-Instruct -cs scores.json \
  -se source.csv -te target.csv -c 1.0
circuitkit heal -m Qwen/Qwen2.5-1.5B-Instruct -p pruned.pt -c scores.pt --task ioi
circuitkit data template rows.csv --clean-prompt "The capital of {country} is" \
  --clean-answer "{capital}" --output task.json
circuitkit data strategies
\end{codeblock}

\medskip
\noindent \texttt{circuitkit run pipeline.yaml} executes an entire discover, evaluate, and intervene run from a version-controlled YAML file, with no Python. The file names the model, task, discovery configuration, the evaluation pillars to compute, the intervention to apply, and the export target:

\begin{codeblock}[listing options={language=yaml}]
# pipeline.yaml
model: meta-llama/Llama-3.2-3B-Instruct
task: boolq

discovery:
  algorithm: eap-ig
  level: neuron
  sparsity: 0.3
  n_examples: 128

evaluate:
  pillars: [patching, ablation, stability]
  n_stability_runs: 3

applications:
  prune:
    sparsity: 0.3
    scope: both

export:
  path: ./output/pruned_checkpoint

benchmark:
  tasks: [boolq]
\end{codeblock}

\subsection{Custom Data End to End}
\label{subsec:customdata-usage}

The three interfaces compose with the custom-data path of Section~\ref{subsec:custom-data} without extra code. A custom task defined by templates over a CSV drops straight into a pipeline, so taking a raw file to a discovered, six-pillar-evaluated circuit, and if desired a pruned checkpoint, needs no bespoke Python. A YAML task definition does the same declaratively, mapping the dataset's columns to prompt and answer roles and optionally naming a corruption strategy when only clean data is present:

\begin{codeblock}[listing options={language=yaml}]
# task.yaml
name: my_task
source:
  type: csv           # csv | jsonl | hf
  path: data.csv
schema:
  prompt: question        # column -> clean prompt
  answer: answer          # column -> clean answer
  corrupted: corrupted            # optional: explicit counter-factual prompt
  corrupted_answer: corrupt_answer
corruption:
  strategy: entity_swap   # only if no explicit corrupted column is given
metric: logit_diff        # logit_diff | kl for discovery; accuracy for reporting
\end{codeblock}

\medskip
\noindent When the CSV already carries an explicit \texttt{corrupted} column, that counterfactual is used directly and no strategy runs; otherwise the named strategy synthesizes it (Section~\ref{subsec:corruption}). The programmatic equivalent, \texttt{Pipeline.from\_custom\_data} with template strings, is the form shown in Section~\ref{subsec:custom-data}.

\section{Stage-Wise Validation and Case Studies}
\label{sec:experiments}

These experiments serve two purposes: E1--E4 validate specific library capabilities under stated configurations, and E5--E7 illustrate how the shared interface supports controlled comparisons between circuit-derived and conventional selectors, reusing runs from our broader actionability audit~\cite{companion2026} that reports the underlying analysis more fully. They are not intended to establish universal rankings between discovery algorithms or intervention strategies. Each study confirms that a stage of the pipeline in Sections~\ref{sec:discovery} through~\ref{sec:applications} behaves as intended on real models, with honest reporting of where circuit guidance does and does not separate from matched baselines. 

Seven studies (E1 through E7) cover discovery validation on a characterized behavior, cross-family coverage, neuron-level granularity, the custom-data path end to end, and one study per primary intervention. Full protocols (models, example counts, sparsity, seeds, and hyperparameters) are collected in Appendix~\ref{app:protocols}; the two cells that remain unmeasured (multi-seed stability outside the stability pillar, and cross-seed intervention reliability) are named as such, not estimated.

Unless a study states otherwise, discovery uses the logit-difference metric at sparsity $\alpha{=}0.3$ and seed 42, faithfulness ratios are reported \emph{signed} (a value below zero marks a behavioral inversion beneath the corrupt-run baseline, not a measurement error; E3 and E4 contain such cells), and downstream accuracy is measured through the \texttt{lm-evaluation-harness} integration of Section~\ref{subsec:benchmarking}. 
The intervention studies (E5--E7) reuse the checkpoints and protocol of our broader actionability audit~\cite{companion2026}, so the library validation and the audit share one set of runs instead of duplicating compute. Appendix~\ref{app:protocols} states every study's parameters in full.

\subsection{Algorithm Validation on IOI (Node Level)}

\paragraph{Question.} Do the stable discovery algorithms recover the \emph{same, correct} circuit for a behavior whose mechanism is already known? (Protocol: Appendix~\ref{app:e1}.)
\label{subsec:e1}
\begin{figure}[htpb]
    \centering
    \includegraphics[width=0.5\linewidth]{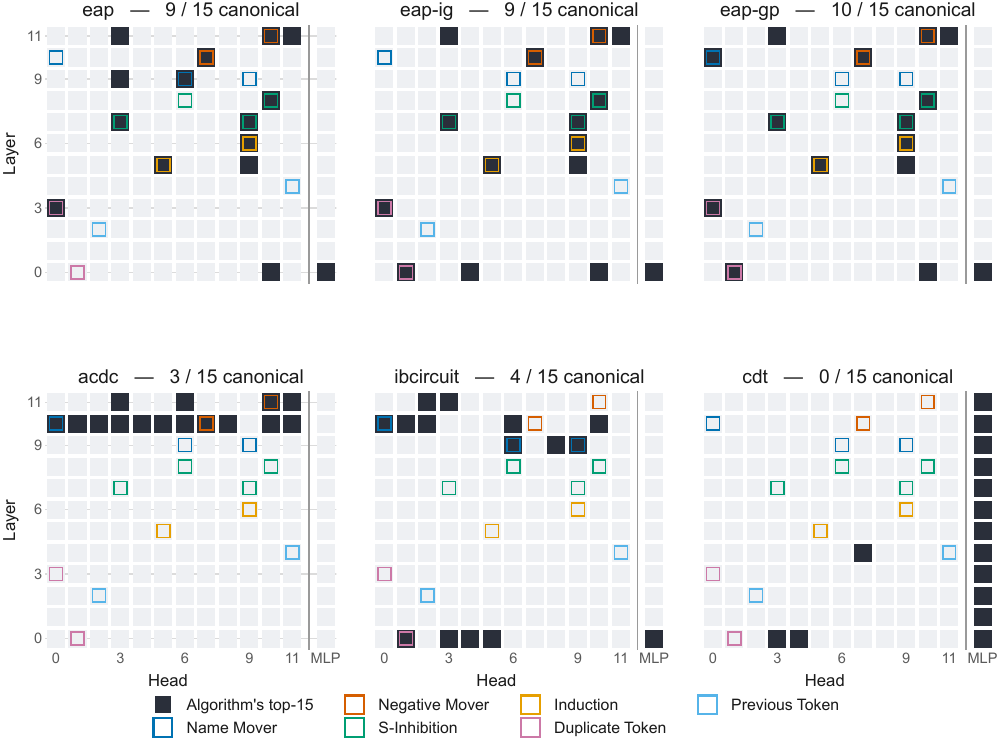}
    \caption{E1: where each algorithm's circuit sits in the model. Each
    panel shows GPT-2 Small's $12{\times}12$ attention-head grid plus a
    right-hand strip for the 12 MLP sublayers. Charcoal squares are the
    algorithm's top-15 nodes; concentric coloured squares mark the 15
    canonical IOI heads by class~\cite{wang2023interpretability}, and
    header counts are canonical heads recovered. The EAP family
    concentrates in mid-to-late layers on the canonical Name-Mover and
    S-Inhibition heads and spends one slot on \texttt{MLP 0}; ACDC's
    tolerance search collects a full row at layer 10 and no MLPs; CD-T
    puts twelve of its fifteen slots on MLPs and only three on attention
    heads, none canonical, which is why its recall is zero on this
    taxonomy.}
    \label{fig:e1-headmap}
    \vspace{-6mm}
\end{figure}

\begin{table}[htpb]
\centering
\caption{E1: six stable algorithms on GPT-2 Small IOI, node level, $\alpha{=}0.3$, seed 42, 256 discovery / 256 evaluation examples. P1 is causal-patching recovery; P2 is ablation faithfulness (the circuit is kept and out-of-circuit components are zero-ablated, Section~\ref{sec:evaluation}); both higher is better. Canonical $J$ / recall compare the discovered top heads against the canonical IOI taxonomy~\cite{wang2023interpretability}. Time is discovery plus two-pillar evaluation wall-clock.}
\vspace{2mm}
\label{tab:e1-ioi}
\footnotesize
\begin{tabular}{@{}lccccc@{}}
\toprule
Algorithm & P1 patching $\uparrow$ & P2 ablation $\uparrow$ & Canonical $J$ $\uparrow$ & Canonical recall $\uparrow$ & Time (s) \\
\midrule
\texttt{eap}       & \textbf{1.000} & 0.883 & 0.429 & 0.600 & 66 \\
\texttt{eap-ig}    & \textbf{1.000} & 0.698 & 0.429 & 0.600 & 87 \\
\texttt{eap-gp}    & \textbf{1.000} & 0.217 & \textbf{0.500} & \textbf{0.667} & 140 \\
\texttt{acdc}      & 0.208 & 0.039 & 0.071 & 0.133 & 3322 \\
\texttt{ibcircuit} & 0.490 & 0.709 & 0.200 & 0.333 & 63 \\
\texttt{cdt}       & 0.894 & \textbf{1.000} & 0.000 & 0.000 & 25 \\
\bottomrule
\end{tabular}
\end{table}

\paragraph{Results.} Table~\ref{tab:e1-ioi} has the per-algorithm scores. The three EAP-family algorithms recover the behavior with perfect patching (P1~$=1.00$) and the highest overlap with the canonical head set (Jaccard 0.43--0.50, recall 0.60--0.67) at minute-scale cost. CD-T is faithful by both pillars (0.89 / 1.00) yet shares \emph{zero} heads with the canonical set, routing twelve of its fifteen slots through MLP sublayers (Figure~\ref{fig:e1-headmap}): a faithful selection whose top components differ from the taxonomy the literature describes. 

IBCircuit lands mid-field (0.49 / 0.71), and ACDC scores poorly under the fixed-sparsity protocol (0.21 / 0.04) at $24$--$133\times$ the cost, a measured limit of projecting its variable-size edge set onto a fixed node budget rather than of the underlying search (Section~\ref{sec:limitations}). Only the overlap columns beside the faithfulness columns tell a canonical circuit apart from an equally faithful non-canonical one, which is the design argument for reporting a panel over a number.

\begin{takeawaybox}\small \textbf{Behavioral recovery and canonical overlap disagree here.} The EAP family recovers the canonical IOI heads at minute-scale cost, while CD-T is equally faithful yet shares zero canonical heads: only the overlap columns beside the faithfulness columns make the distinction visible (Table~\ref{tab:e1-ioi}, Figure~\ref{fig:e1-headmap}).
\end{takeawaybox}

\subsection{Cross-Family Discovery}
\label{subsec:e2}

\paragraph{Question.} Does discovery run, and produce faithful, stable circuits, across model families of different architecture and scale? (Protocol: Appendix~\ref{app:e2}.)
\begin{table}[htpb]
\vspace{-5mm}
\centering
\caption{E2: cross-family discovery with \texttt{eap-ig}, node level, $\alpha{=}0.3$, seed 42, two tasks per family. P1 patching, P2 ablation (circuit-only, zero replacement), P5 baselines pillar (improvement over a size-matched random subgraph), P3 stability (mean pairwise Jaccard, seeds 42--44). $\dagger$: P5 ratio undefined because the random subgraph's own metric is $\leq 0$ while the circuit's is positive: the strongest separation, reported as a status flag rather than a ratio.}
\vspace{2mm}
\label{tab:e2-crossmodel}
\footnotesize
\setlength{\tabcolsep}{4pt}
\begin{tabular}{@{}lcccc c cccc@{}}
\toprule
 & \multicolumn{4}{c}{IOI} & & \multicolumn{4}{c}{Greater-Than} \\
\cmidrule(lr){2-5}\cmidrule(lr){7-10}
Model (size) & P1 & P2 & P5 & P3 $J$ & & P1 & P2 & P5 & P3 $J$ \\
\midrule
GPT-2 Small (124M)   & 1.000 & 0.698 & 2.1$\times$ & 0.803 & & 0.973 & 0.700 & $\dagger$ & 0.919 \\
Pythia-1.4B          & 0.914 & 0.542 & 3.4$\times$ & 0.896 & & 0.978 & 0.805 & $\dagger$ & 0.869 \\
Llama-3.2-1B         & 1.000 & 0.346 & 7.2$\times$ & 0.833 & & 1.000 & 0.654 & $\dagger$ & 0.801 \\
Gemma-2-2B           & 1.000 & 0.244 & 1.2$\times$ & 0.812 & & 0.951 & 0.858 & $\dagger$ & 0.874 \\
Qwen2.5-1.5B         & 1.000 & 0.799 & $\dagger$   & 0.856 & & 0.995 & 0.961 & $\dagger$ & 0.883 \\
Phi-2 (2.8B)         & 0.977 & 1.000 & 3.2$\times$ & 0.879 & & 1.000 & 0.907 & $\dagger$ & 0.891 \\
\bottomrule
\end{tabular}
\end{table}

\paragraph{Results.} All twelve cells completed (Table~\ref{tab:e2-crossmodel}) with no code change, only a model name. Patching recovery is high everywhere (P1~$\geq 0.91$, six cells at exactly $1.00$) and stability is strong across all twelve cells ($J = 0.80$--$0.92$); every circuit beats a size-matched random subgraph, by $1.2\times$ to $7.2\times$ where the ratio is defined and by status flag in the seven $\dagger$ cells where the random subgraph's own metric is $\leq 0$. 
Ablation sufficiency is where the spread lives: P2 ranges from 0.24 (Gemma-2-2B\,/\,IOI) to 1.00 (Phi-2\,/\,IOI) across cells whose patching scores are nearly indistinguishable; this spread reflects the documented dependence of hard ablation on the replacement distribution~\cite{miller2024faithfulness}. 
A reader who took P1 alone would call the twelve circuits interchangeable; one who took P2 alone would call discovery a failure on Gemma. The disagreement between the columns is the finding.

\begin{takeawaybox}\small \textbf{EAP-IG runs across six families unchanged.} Discovery runs faithfully and stably across six families and two tasks with only a model-name change; ablation sufficiency, not patching, is where the circuits differ (Table~\ref{tab:e2-crossmodel}).
\end{takeawaybox}

\FloatBarrier
\subsection{Neuron-Level Discovery}
\label{subsec:e3}

\paragraph{Question.} Does discovery at neuron granularity produce a faithful, stable circuit, and how does it compare to the node-level circuit for the same behavior? (Protocol: Appendix~\ref{app:e3}.)
\begin{table}[htpb]
\centering
\caption{E3: neuron-level IOI discovery on Llama-3.2-1B, $\alpha{=}0.3$, seed 42, MLP scored at \texttt{post\_act}; node-level reference in the last block. P2 is signed: the negative EAP cell marks a behavioral inversion under hard zero-ablation (see text).}
\vspace{2mm}
\label{tab:e3-neuron}
\footnotesize
\begin{tabular}{@{}llccc c@{}}
\toprule
Level & Method & P1 $\uparrow$ & P2 $\uparrow$ & Units kept (of total) & Disc.\ (s) \\
\midrule
Neuron & \texttt{eap}    & 0.893 & $-0.350$ & 825{,}754 / 1{,}179{,}648 & 77 \\
Neuron & \texttt{eap-ig} & \textbf{1.000} & 0.241 & 825{,}754 / 1{,}179{,}648 & 147 \\
\midrule
Node   & \texttt{eap}    & 0.958 & 0.215 & 371 / 528 & 82 \\
Node   & \texttt{eap-ig} & \textbf{1.000} & 0.346 & 371 / 528 & 140 \\
\bottomrule
\end{tabular}
\end{table}

\paragraph{Results.} Table~\ref{tab:e3-neuron} collects both granularities. EAP-IG preserves perfect patching at neuron level (P1~$=1.00$, as at node level) with circuit-only sufficiency P2~$=0.24$ against 0.35 at node level, at essentially the same discovery cost: finer granularity holds faithfulness at fixed sparsity while exposing exactly the units the intervention modules act on. Single-point-gradient EAP degrades more visibly (P1~$=0.89$, signed P2~$=-0.35$): under hard zero-ablation the isolated neuron-level circuit is driven \emph{below} the corrupt-run baseline (raw logit difference $-0.57$ against a clean-run $1.69$), so a bounded score would clamp the cell to $0.00$ and hide the inversion, which is why we report the signed ratio (Section~\ref{sec:evaluation}). Integrated-gradient interpolation thus matters most where single-point gradients are noisiest, that is, at fine granularity; at $\alpha{=}0.3$ both granularities retain 70\% of units (371 of 528 nodes, and 825{,}754 of 1{,}179{,}648 neurons).

\begin{takeawaybox}\small \textbf{Neuron-level discovery preserves patching here.} EAP-IG holds perfect patching at neuron granularity for essentially the same cost as node level, exposing the exact units interventions act on, while single-point EAP inverts under hard ablation (Table~\ref{tab:e3-neuron}).
\end{takeawaybox}

\FloatBarrier
\subsection{The Custom-Data Path End to End}
\label{subsec:e4}

\paragraph{Question.} Does the template-driven custom-data path carry a user-defined, non-canonical dataset through discovery and evaluation, on both the paired and the clean-only routes? (Protocol, including the refusal-circuit metric convention, in Appendix~\ref{app:e4}.)
\begin{table}[!ht]
\centering
\caption{E4: custom-data jailbreak task on Qwen2.5-1.5B-Instruct, $\alpha{=}0.3$, seed 42 (multi-seed stability pending). The negative P2 marks a behavioral inversion under hard ablation, reported signed rather than clamped to zero (see text). $^{\dagger}$Clean-only sufficiency: circuit-only vs.\ full-model P(correct).}
\vspace{2mm}
\label{tab:e4-custom}
\footnotesize
\setlength{\tabcolsep}{4pt}
\begin{tabular}{@{}lcc@{}}
\toprule
Quantity & Paired (\texttt{eap-ig}, node) & Clean-only (\texttt{ibcircuit}, neuron) \\
\midrule
Records kept (of 334)          & 265 & 334 \\
Circuit size at $\alpha{=}0.3$ & 256 / 364 nodes & 30{,}106 / 43{,}008 neurons \\
P1: patching recovery          & 0.854 & 0.902 vs.\ 0.874$^{\dagger}$ \\
P2: ablation (signed)          & $-2.61$ (raw $-1.55$) & --- \\
P5: vs.\ random / magnitude    & $1.71\times$ / $1.44\times$ & --- \\
P4: answer-preserving $\Delta$ & 0.00 (paraphrase) & --- \\
Runtime (discovery + eval.)    & 4.5 min & 2.7 min \\
\bottomrule
\end{tabular}
\end{table}

\begin{figure}[!htbp]
    \centering
    \includegraphics[width=0.6\linewidth]{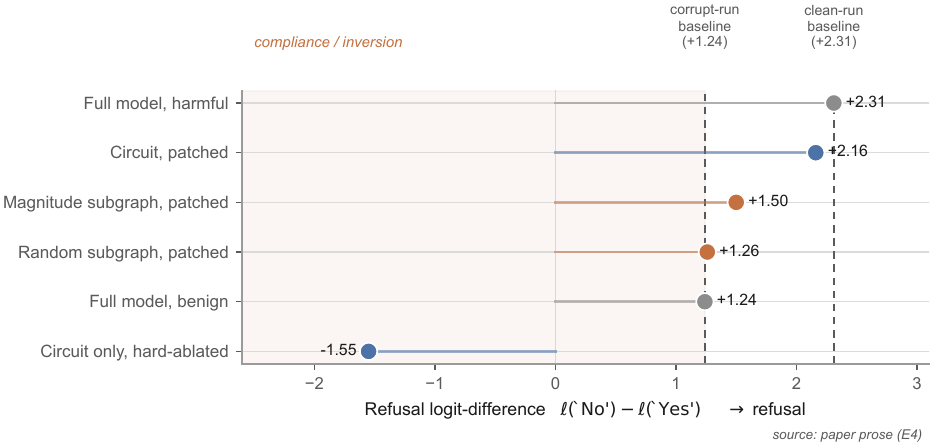}
    \caption{E4: refusal metric on the paired-route circuit under six
    conditions. Full-model harmful and benign requests set the dashed
    reference lines; the shaded region below the benign baseline is the
    compliance/inversion band. The patched circuit sits close to the
    harmful baseline (85\% recovery); the size-matched random and
    magnitude subgraphs sit near the benign baseline; and the hard-ablated
    circuit falls below both, at $-1.55$, i.e.\ isolated from its
    surrounding computation the truncated model's binary refusal proxy inverts toward compliance. A
    bounded faithfulness score would clip the last row at zero and hide
    the inversion.}
\vspace{2mm}
    \label{fig:e4-refusal}
\end{figure}

\paragraph{Results.} Both routes complete without hand-written pairing code (Table~\ref{tab:e4-custom}); the alignment filter kept 265 of 334 rows, dropping 69 misaligned pairs rather than padding them. The paired circuit recovers 85\% of the full-model refusal gap under patching and beats size-matched baselines ($1.71\times$ random, $1.44\times$ magnitude), yet when Pillar 2 isolates it under hard zero-ablation the refusal metric on harmful requests falls to $-1.55$, below the benign-request baseline (signed ratio $-2.61$): stripped of its surrounding computation, the truncated model \emph{prefers to comply} (Figure~\ref{fig:e4-refusal}). Pillar 4 reports the circuit unchanged under answer-preserving paraphrase ($\Delta{=}0.00$) and near-unchanged under role-swap ($\Delta{=}0.05$), with length-changing variants refused by the alignment guard; on the clean-only route, IBCircuit's neuron-level circuit alone matches the full model (P(correct) 0.90 vs.\ 0.87) despite never seeing a corrupt pair. The systems claim is that a raw CSV reached a multi-pillar-evaluated circuit down two algorithmic routes with zero pairing code; the scientific one is that localizing a safety behavior and \emph{preserving} it under intervention are different achievements, which bears directly on work that localizes and edits refusal through a few directions or components~\cite{arditi2024refusal, kasliwal2026cdeltatheta}. Cross-route component agreement and multi-seed stability for this cell were not run and are not claimed.

\begin{takeawaybox}\small \textbf{Soft patching and hard ablation disagree here.} A raw CSV reaches a multi-pillar-evaluated refusal circuit down both routes with zero pairing code, and the same circuit that recovers 85\% of the refusal gap under the soft counterfactual inverts toward compliance under the hard one (Table~\ref{tab:e4-custom}, Figure~\ref{fig:e4-refusal}).
\end{takeawaybox}

\FloatBarrier

\subsection{Circuit-Guided Structural Pruning}
\label{subsec:e5}

\paragraph{Question.} Does a circuit-guided importance signal prune competitively at matched sparsity, and does a circuit's faithfulness predict how safe its ranking is to act on? (Protocol: Appendix~\ref{app:e5}.)
\begin{table}[!ht]
\centering
\caption{E5: circuit-guided vs.\ standard structural pruning on Llama-3.2-3B-Instruct / BoolQ at 30\% sparsity, seed 42. accRet = post-pruning accuracy / base accuracy (base 0.755); PPL = WikiText-2 perplexity of the pruned model (unpruned: 11.05). Magnitude, Taylor, and Wanda are deterministic weight-saliency baselines.}
\vspace{2mm}
\label{tab:e5-pruning}
\footnotesize
\begin{tabular}{@{}llcc@{}}
\toprule
Selector & Family & accRet $\uparrow$ & WikiText-2 PPL $\downarrow$ \\
\midrule
Taylor            & saliency & \textbf{1.027} & 93.6 \\
\texttt{ibcircuit} & circuit  & 0.986 & 156.2 \\
\texttt{eap-gp}    & circuit  & 0.927 & 56.0 \\
\texttt{eap}       & circuit  & 0.866 & 90.0 \\
\texttt{eap-ig}    & circuit  & 0.571 & \textbf{35.5} \\
Random            & control  & 0.519 & $3.4 \times 10^{5}$ \\
\texttt{cdt}       & circuit  & 0.502 & 1406.6 \\
Magnitude         & saliency & 0.501 & 3335.1 \\
Wanda             & saliency & 0.501 & $1.3 \times 10^{4}$ \\
\bottomrule
\end{tabular}
\end{table}

\begin{figure}[!htbp]
    \centering
    \includegraphics[width=0.6\linewidth]{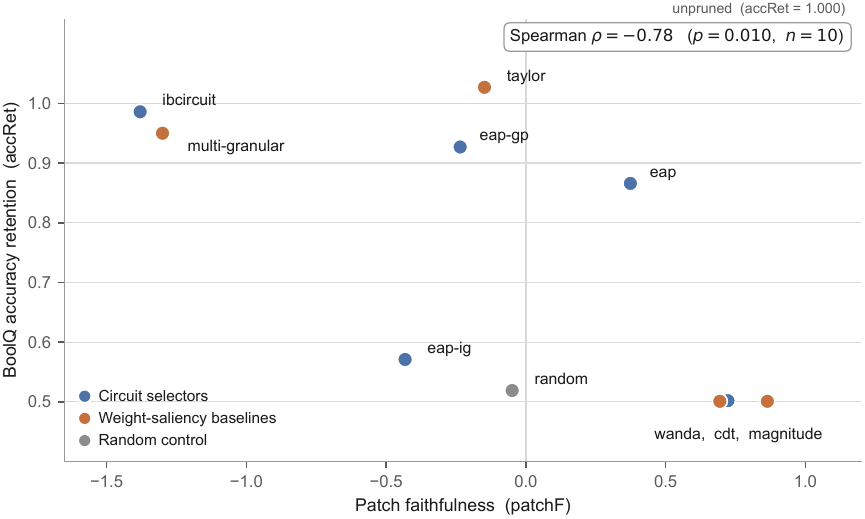}
    \caption{E5: patch faithfulness against pruning retention over the
    audit's ten selectors on this cell ($\rho = -0.78$, $p = 0.010$;
    Llama-3.2-3B-Instruct\,$\cdot$\,BoolQ, 30\% sparsity, seed 42). On
    this cell the more patch-faithful a selector's circuit, the worse it
    prunes; \texttt{ibcircuit} sits at the
    faithful-and-effective corner while \texttt{cdt}, \texttt{wanda}, and
    \texttt{magnitude} cluster at the faithful-and-degenerate corner.
    Values from our audit table.}
\vspace{2mm}
    \label{fig:e5-scatter}
\end{figure}

\paragraph{Results.} Circuit-guided pruning is competitive at the top (Table~\ref{tab:e5-pruning}): IBCircuit retains 98.6\% of base accuracy, within 4.1\,pp of the strongest overall criterion (Taylor, 102.7\%), with EAP-GP at 92.7\% and EAP at 86.6\%, all far above magnitude, Wanda, and random (50.1--51.9\%), whose pruned models collapse to degenerate outputs (WikiText-2 PPL $\geq 3{,}000$). But the spread \emph{within} the circuit family is wide: EAP-IG (57.1\%) and CD-T (50.2\%) rank among the worst compressors despite strong patch faithfulness, and across the audit's ten-selector grid patch faithfulness \emph{anti-correlates} with retention here (Spearman $\rho = -0.78$, $p = 0.010$; Figure~\ref{fig:e5-scatter}). The perplexity column tracks a separate axis: EAP-IG prunes to the lowest perplexity of any selector (35.5) while retaining barely half IBCircuit's task accuracy. ``Circuit-guided'' names where the signal came from, not how well it ranks components for removal, so validating the selector extrinsically through \texttt{ck.benchmark} before committing to a cut is, on this evidence, not optional.

\begin{takeawaybox}\small \textbf{Patch faithfulness does not predict pruning here.} Circuit importance prunes competitively at the top, but which discovery algorithm produces it matters more than its faithfulness score, which anti-correlates with retention on this cell ($\rho = -0.78$; Table~\ref{tab:e5-pruning}, Figure~\ref{fig:e5-scatter}).
\end{takeawaybox}

\FloatBarrier

\subsection{Circuit-Guided Mixed-Precision Quantization}
\label{subsec:e6}

\paragraph{Question.} Does circuit importance choose \emph{which} components to protect under a fixed precision budget better than the alternatives, or is quantization insensitive to the choice? (Protocol: Appendix~\ref{app:e6}.)
\begin{table}[!ht]
\centering
\caption{E6: mixed-precision quantization (3-bit weights / 8-bit activations, 5\% floating-point fallback) at matched budget; Spearman correlations over $n{=}9$ selector-means (45 cells, two models).}
\vspace{2mm}
\label{tab:e6-quant}
\footnotesize
\begin{tabular}{@{}lcc@{}}
\toprule
Correlation (quantization) & $\rho$ & $p$ \\
\midrule
$\rho(\text{patchF}, \text{accRet})$ & $+0.23$ & 0.55 \\
$\rho(\text{ablF}, \text{accRet})$   & $+0.73$ & 0.031 \\
$\rho(\text{patchF}, \text{ablF})$   & $+0.533$ & 0.139 \\
\bottomrule
\end{tabular}
\end{table}

\paragraph{Results.} Unlike pruning, quantization retention is largely insensitive to the selector's patch faithfulness (Table~\ref{tab:e6-quant}): the correlation is non-significant ($\rho = +0.23$, $p = 0.55$, $n = 9$ selector-means). Ablation faithfulness, by contrast, does predict retention ($\rho = +0.73$, $p = 0.031$), consistent with its zero-replacement counterfactual being closer to what quantization error inflicts than patchF's activation-swap. The patchF null is the \emph{predicted} behavior: quantization distributes precision across all components rather than forcing a hard keep/discard partition, so the ranking errors that punish pruning have no channel through which to bite; the two faithfulness pillars therefore agree here ($\rho = +0.53$) though they decouple under pruning ($-0.22$). Two consequences follow: circuit-guided tier selection did not significantly underperform uniform assignment at matched budget in this setting, and compression decisions should be validated per intervention type.

\begin{takeawaybox}\small \textbf{Little selector sensitivity under this quantization recipe.} Under a fixed precision budget, patch faithfulness does not predict retention while ablation faithfulness does, so circuit-guided tier selection did not underperform uniform assignment in this setting (Table~\ref{tab:e6-quant}).
\end{takeawaybox}

\FloatBarrier
\subsection{Circuit-Guided Selective Fine-Tuning}
\label{subsec:e7}

\paragraph{Question.} Does fine-tuning only the circuit's most important parameters improve a task more than a random selection of equal budget, and what does the restriction do to the rest of the model? (Protocol: Appendix~\ref{app:e7}.)
\begin{table}[!ht]
\centering
\caption{E7: circuit-guided vs.\ random-budget vs.\ full-parameter selective fine-tuning on Llama-3.2-3B-Instruct / BoolQ (top-30\% static gradient mask, 3 epochs, 500 train / 300 eval examples, seed 42; pre-FT accuracy 0.413 under the fine-tuning eval protocol). All conditions share data and hyperparameters and differ only in which weights receive gradients; random and full-parameter conditions are trained once per run group, hence repeated values. PPL = post-training WikiText-2 perplexity (unmodified model: 11.05).}
\vspace{2mm}
\label{tab:e7-finetune}
\footnotesize
\setlength{\tabcolsep}{4pt}
\begin{tabular}{@{}lcccccc@{}}
\toprule
 & \multicolumn{3}{c}{BoolQ accuracy $\uparrow$} &
   \multicolumn{3}{c}{WikiText-2 PPL $\downarrow$} \\
\cmidrule(lr){2-4}\cmidrule(l){5-7}
Selector & Circuit & Random & Full & Circuit & Random & Full \\
\midrule
\texttt{ibcircuit} & \textbf{0.697} & 0.667 & 0.677 & 12.1 & 11.7 & 14.2 \\
\texttt{eap-ig}    & 0.670 & 0.663 & 0.663 & \textbf{12.0} & 13.9 & 20.6 \\
\texttt{eap}       & 0.653 & 0.657 & 0.683 & 14.1 & 13.8 & 22.0 \\
Taylor             & 0.690 & 0.663 & 0.663 & \textbf{11.6} & 13.9 & 20.6 \\
\bottomrule
\end{tabular}
\end{table}

\paragraph{Results.} On task accuracy, circuit selection does not separate from random selection at matched budget (Table~\ref{tab:e7-finetune}): the best single cell is $+3.0$\,pp (IBCircuit), and pooling the audit's 16 (selector, model) cells the mean $\Delta(\text{circuit} - \text{random})$ is $-0.001$, a null result we report as such. The coherence column separates sharply, but not along the circuit axis: masked fine-tuning of either kind leaves WikiText-2 perplexity near the unmodified model's 11.05 (12.0--14.1 for circuit masks, 11.7--13.9 for random masks of the same budget), while full-parameter training on the same 500 examples degrades it to 14.2--22.0. What protects general language modeling is therefore \emph{the budget constraint itself}: freezing 70\% of the model bounds how far the update can drift, and any 30\% mask inherits that bound. Selective fine-tuning through \CK{} is the right tool for a targeted update that leaves the rest of the model intact, but circuit-derived importance is not what earns that property, so we report the study as a null on both axes.

\begin{takeawaybox}\small \textbf{Mask identity does not beat the budget-matched control.} Circuit selection does not beat a random mask at matched budget; freezing 70\% of the model, not which 30\% is chosen, is what protects coherence (Table~\ref{tab:e7-finetune}).
\end{takeawaybox}

\FloatBarrier
\subsection{Reading the Intervention Studies Together}
\label{subsec:e-stability}

Read together, E5 through E7 show one pattern: \emph{the value of a circuit signal depends on the intervention that consumes it}. Pruning forces a hard keep/discard partition, so a ranking error is unrecoverable and selector quality is decisive (E5). Quantization forces no partition, so patchF ranking errors have no channel through which to bite, while ablation faithfulness, whose counterfactual is closer to quantization error, does predict retention (E6). Selective fine-tuning constrains a budget rather than ranking within it, so what matters is that 70\% of the model is frozen, not which 30\% was chosen (E7).

Three interventions, three different relationships to the same signal: that is the toolkit-level case for a faithfulness \emph{panel} over a single score, and for closing the loop extrinsically. Across our broader audit grid~\cite{companion2026}, run entirely through \CK{}'s selector registry, evaluation pillars, and benchmarking layer, patch faithfulness correlates significantly \emph{negatively} with pruning retention on the sharpest cell and changes sign across tasks. A faithfulness score is evidence about what a circuit explains, not a forecast of what will happen when you act on it; downstream effects should therefore be measured directly rather than inferred from intrinsic faithfulness alone, which is what \texttt{ck.benchmark} provides. Relating discovery stability $J$ to cross-seed intervention reliability ($\text{SD}_{\text{rel}}$, Pillar 7) requires multi-seed re-discovery per intervention; with the intervention grid single-seed, we leave it to future work.

\section{Conclusion}
\label{sec:conclusion}

\emph{CircuitKIT} provides common infrastructure for moving from circuit discovery to diagnostic evaluation, downstream application, and standardized benchmarking. Its central contribution is an interface that allows methods with different implementations to produce and consume a shared circuit representation, together with declarative task adapters for structured user data. Our studies validate these pathways across discovery algorithms, model families, custom-data routes, and downstream applications. They also illustrate why \emph{CircuitKIT} reports multiple diagnostics and downstream measurements separately: behavioral recovery, component overlap, ablation sufficiency, and application performance need not agree. \emph{CircuitKIT} therefore standardizes how circuit hypotheses are produced, evaluated, and applied without assuming that any discovery method or intrinsic score is universally reliable. Released source-available with extensible registries for its algorithms, selectors, corruption strategies, and model families, and with sparse-autoencoder features~\cite{marks2025sparse} a natural next granularity, \emph{CircuitKIT} is built to grow with the field.

The library is available at \url{https://github.com/Lexsi-Labs/CircuitKIT}.

\section{Limitations, Ethics, and Broader Impact}
\label{sec:limitations}
\label{sec:ethics}

\emph{CircuitKIT} has several limitations. Discovery and evaluation run on models supported by the pinned TransformerLens version, but the intervention modules require an architecture-registry entry, so intervention coverage trails discovery and inherits TransformerLens's architecture catalogue and version pins .

Individual methods carry their own costs: ROME/MEMIT editing needs a model- and layer-specific key covariance, which \emph{CircuitKIT} reduces to one cached \texttt{get\_covariance} call but still requires a user-supplied corpus; IBCircuit trains its mask in a single batch, so its activation footprint grows with model size (hence the GPT-2-scale six-algorithm comparison in E1); and ACDC is node-only by construction, since its greedy edge search has no per-channel analogue.

On the data side, the corruption strategies (Section~\ref{subsec:corruption}) assume identifiable syntactic structure and may find nothing to change on instruction-tuned or safety-relevant prompts, where user-supplied corrupt columns are the recommended route and confirming that auto-synthesized pairs carry meaningful signal on datasets such as AdvBench~\cite{zou2023advbench}, TruthfulQA~\cite{lin-etal-2022-truthfulqa}, and StereoSet~\cite{nadeem-etal-2021-stereoset} remains future work. 

Finally, statistical power is bounded: the stability pillar uses only three data resamples (seeds 42--44), the other studies are single-seed (so cross-seed reliability, $\text{SD}_{\text{rel}}$, Pillar 7, is not yet reported and the stability-to-reliability association of Section~\ref{subsec:e-stability} is unmeasured), and E5--E7 report one Llama-3.2-3B-Instruct\,$\cdot$\,BoolQ cell of a broader audit~\cite{companion2026} that replicates qualitatively on Gemma-3-4B-IT and a second task, so the specific numbers, the $\rho = -0.78$ pruning anti-correlation above all, are one cell rather than a population estimate.

\emph{CircuitKIT} targets AI-safety and efficiency uses: refusal and bias auditing, factual editing, unlearning evaluation, and circuit-guided compression, part of a broader case for treating interpretability as a design principle for alignment and governance rather than post-hoc explanation~\cite{sengupta2025interpalign}. The dual-use risk is concrete rather than abstract: circuit localization is symmetric, and E4 (Section~\ref{subsec:e4}) shows a discovered refusal circuit that, isolated under hard ablation, inverts from refusal toward compliance; we report it because the same measurement is what would warn a practitioner that an intervention had crossed a line. The release mitigates this without pretending to eliminate it: the library ships aggregate scores and evaluation reports, not harmful prompts or pre-ablated checkpoints; the steering, editing, and fine-tuning modules require explicit import; and the case-study data is research-use only. 

\emph{CircuitKIT} is released under the Lexsi Labs Source Available License (LSAL~v1.1), which grants free use for research, evaluation, education, and audit while prohibiting commercial exploitation and uses that degrade deployed models' safety behaviors; it is deliberately \emph{not} an OSI-approved open-source license, since that restriction is the mechanism bounding the dual-use risk. Commercial licensing is available separately; all benchmark datasets used here are public and cited, and exported checkpoints inherit their base models' licenses.


\bibliographystyle{unsrt}
\bibliography{references}
\clearpage
\appendix
\section{Experimental Protocols}
\label{app:protocols}

This appendix collects the full protocol for each study of Section~\ref{sec:experiments}: models, example counts, sparsity, seeds, and hyperparameters. The tables, figures, and results remain in the main text; only the reproduction detail lives here.

\subsection{Shared Setup}
\label{app:common-setup}

Unless a study states otherwise, discovery uses the logit-difference metric with 256 discovery / 256 evaluation examples at sparsity $\alpha{=}0.3$ and seed 42, the stability pillar (P3) re-discovers the circuit over three independent data resamples (seeds 42--44) and reports the mean pairwise Jaccard $J$ (Equation~\ref{eq:jaccard}), and downstream accuracy is measured through the \texttt{lm-evaluation-harness} integration of Section~\ref{subsec:benchmarking}. Faithfulness ratios are reported \emph{signed}: a ratio falls below zero when an intervention drives the task metric beneath the corrupt-run baseline, and we report that value instead of clamping it to zero, because the sign carries information (E3 and E4 contain such cells and discuss them). The intervention studies (E5--E7) reuse the runs and protocol of our broader actionability audit~\cite{companion2026} (128 discovery examples disjoint from evaluation, patching faithfulness on a 300-example held-out partition, bfloat16, seed 42), so the library validation and the audit share one set of checkpoints instead of duplicating compute. Scripts and configurations for every study are released with the library.

\subsection{IOI Validation}
\label{app:e1}

IOI is the most-replicated task in the circuit-discovery literature~\cite{wang2023interpretability, conmy2023towards, hanna2024faith}, and its circuit is characterized head by head, which makes it the natural cell for validating that our implementations find the right components and not merely a high-scoring subgraph. We run all six stable algorithms (\texttt{eap}, \texttt{eap-ig}, \texttt{eap-gp}, \texttt{acdc}, \texttt{ibcircuit}, \texttt{cdt}) on GPT-2 Small at \emph{node} level, $\alpha{=}0.3$, seed 42, with 256 discovery and 256 evaluation examples from the built-in \texttt{ioi} task. We run at node level because it lets us compare the discovered heads directly against the canonical IOI taxonomy (Name-Mover, Backup Name-Mover, S-Inhibition, Induction, Duplicate-Token, and Previous-Token heads). For each algorithm we report the causal-patching (P1) and ablation (P2) scores, discovery wall-clock time, and the overlap between the top-scoring heads and the canonical set. GPT-2 Small is used because it is the model on which the IOI circuit was originally established and the one scale at which all six algorithms, including the memory-heavy IBCircuit, fit comfortably.

\subsection{Cross-Family Discovery}
\label{app:e2}

Holding the algorithm fixed at \texttt{eap-ig} (node level, $\alpha{=}0.3$, stability over data-resample seeds 42--44), we run discovery on six families spanning the architectural variation TransformerLens supports: GPT-2 Small (124M, learned positional embeddings), Pythia-1.4B (parallel attention/MLP), Llama-3.2-1B (RoPE, grouped-query attention), Gemma-2-2B (RoPE, gated MLP, logit soft-capping), Qwen2.5-1.5B (RoPE, GQA, untied embeddings), and Phi-2 (2.8B; parallel attention/MLP with partial-rotary embeddings). Each model runs on two tasks: \texttt{ioi}, as a mechanism whose circuit is architecture-independent in principle, and \texttt{greater\_than}~\cite{hanna2023greaterthan}, the numerical year-completion behavior, which exercises a different mechanism and confirms the pipeline is not IOI-specific. We report P1 (patching), P2 (ablation), P5 (the baselines pillar: random and size-matched magnitude subgraphs), and P3 (stability $J$ over seeds 42--44) for each of the twelve (model, task) cells. This is the study that substantiates the model-coverage claim of Section~\ref{subsec:coverage}. Discovery wall-clock spans 86\,s (GPT-2, Greater-Than) to 366\,s (Phi-2, IOI) per run at 256 examples, scaling with parameter count and with no family-specific tuning; only IBCircuit-style batch capping is architecture-sensitive (Section~\ref{sec:limitations}).

\subsection{Neuron-Level Discovery}
\label{app:e3}

On a single model and task (Llama-3.2-1B, \texttt{ioi}), we run \texttt{eap} and \texttt{eap-ig} at \emph{neuron} level ($\alpha{=}0.3$, seed 42, 256 discovery / 256 evaluation examples), scoring attention at per-head channels and the MLP at its post-activation hidden units (\texttt{mlp\_hook="post\_act"}, Section~\ref{subsec:granularity}). We report P1 and P2, the unit count retained at the target sparsity, and, for context, the same cell at node level; multi-seed stability for this cell is pending. This study substantiates that neuron level is a first-class granularity rather than a coarse-to-fine afterthought, which matters because the intervention modules act on exactly these neuron indices.

\subsection{Custom-Data Path}
\label{app:e4}

We use a binary jailbreak-detection dataset (a 334-row CSV of benign requests, harmful requests, and yes/no answers) that ships with the library, and take it end to end through \texttt{Pipeline.from\_custom\_data} on Qwen2.5-1.5B-Instruct. Two routes are run from the \emph{same} CSV: the paired route (clean and corrupt templates supplied, \texttt{eap-ig}, node level) and the clean-only route (corrupt templates omitted, \texttt{ibcircuit}, neuron level), both at $\alpha{=}0.3$, seed 42. One design choice governs how the numbers must be read: because the target is the \emph{refusal} circuit rather than the compliance circuit, the \emph{harmful} request is assigned the clean role (it elicits the behavior under study, answering ``No''), and the benign request supplies the corrupt counterpart (``Yes''). The task metric is therefore the refusal logit difference, $\mathrm{logit}(\text{`No'}) - \mathrm{logit}(\text{`Yes'})$, and a \emph{negative} score is a behavioral inversion, not an error. For the paired route we report patching, ablation, baselines, and robustness; for the clean-only route, the circuit-only sufficiency check. The two routes exercise the two halves of the custom-data contract of Section~\ref{subsec:custom-data}: template pairing with token alignment for the contrastive algorithm, and clean-only routing for the trainable one.

\subsection{Structural Pruning}
\label{app:e5}

On Llama-3.2-3B-Instruct / BoolQ we prune 30\% of attention heads and MLP blocks (structured zero-masking, the same operation \texttt{ck.prune} applies; Section~\ref{subsec:pruning}) using each importance selector in the registry of Section~\ref{sec:extensibility}, everything else held constant: the five circuit-discovery selectors (\texttt{eap}, \texttt{eap-ig}, \texttt{eap-gp}, \texttt{ibcircuit}, \texttt{cdt}) against the weight-saliency baselines (magnitude, Taylor, Wanda) and a random control. Discovery uses 128 examples disjoint from evaluation; accuracy is measured through the harness (base accuracy 0.755) and reported as retention $\mathrm{accRet} = \mathrm{acc}_{\mathrm{pruned}}/\mathrm{acc}_{\mathrm{base}}$; WikiText-2 perplexity tracks general-LM coherence. These runs come from our broader actionability audit~\cite{companion2026} spanning ten selectors, two models, and two tasks; here we report its Llama\,$\cdot$\,BoolQ cell as the pruning validation study.

\subsection{Mixed-Precision Quantization}
\label{app:e6}

Reusing the E5 circuits and selectors on Llama-3.2-3B-Instruct, each selector assigns which components keep higher precision under one fixed mixed-precision recipe (3-bit weights, 8-bit activations, and a 5\% floating-point fallback) at matched budget across all rows, so the comparison isolates \emph{which} components are protected (Section~\ref{subsec:quantization}). The grid spans nine selectors on two models (45 cells); the downstream metric is accuracy retention through the harness.

\subsection{Selective Fine-Tuning}
\label{app:e7}

On Llama-3.2-3B-Instruct / BoolQ, each selector's top-30\% components are resolved to weight-matrix indices and trained through a static gradient mask (Section~\ref{subsec:finetuning}) for 3 epochs on 500 BoolQ training examples (300 held-out evaluation examples, lr $2{\times}10^{-5}$, bfloat16, seed 42), starting from a pre-fine-tuning accuracy of 0.413. The conditions, identical in every respect except \emph{which} parameters receive gradients, are: the circuit mask, a random mask at the same 30\% budget, and unrestricted full-parameter training. Post-training checkpoints are scored for BoolQ accuracy and WikiText-2 perplexity; the grid spans eight selectors on two models (48 runs). The Gemma-3-4B-IT replication is where full-parameter training on 500 examples is itself unstable, confounding that comparison (Section~\ref{sec:limitations}).

\end{document}